\pdfoutput=1

\documentclass[11pt]{article}

\usepackage[final]{acl}

\usepackage{times}
\usepackage{latexsym}

\usepackage[T1]{fontenc}

\usepackage[utf8]{inputenc}

\usepackage{microtype}

\usepackage{inconsolata}

\usepackage{graphicx}
\usepackage{amsmath}
\usepackage{booktabs}
\usepackage{shadowtext}

\usepackage{xcolor}
\usepackage{graphicx}
\usepackage{enumitem}
\usepackage{rotating}
\usepackage{fancyvrb}
\usepackage{xparse}
\usepackage{pifont}

\newcommand{\noto}[0]{\textsc{noto}\,}

\title{
    {\it None of the Others}: a General Technique to Distinguish Reasoning from Memorization in Multiple-Choice LLM Evaluation Benchmarks
    
    }

\author{Eva Sánchez Salido, Julio Gonzalo, Guillermo Marco \\
UNED Research Center in Natural Language Processing and Information Retrieval\thanks{\href{https://sites.google.com/view/nlp-uned/home}{nlp.uned.es}}\\
ETSI Informática, UNED - Juan del Rosal, 16 28040 Madrid, Spain\\
\small{\textbf{Correspondence:} \href{mailto:julio@lsi.uned.es}{julio@lsi.uned.es}}
}

\begin{document}

\maketitle

\begin{abstract}
In LLM evaluations, reasoning is often distinguished from recall/memorization by performing numerical variations to math-oriented questions. Here we introduce a general variation method for multiple-choice questions that completely dissociates the correct answer from previously seen tokens or concepts, requiring LLMs to understand and reason (rather than memorizing) in order to answer correctly.  
Using this method, we evaluate state-of-the-art proprietary and open-source LLMs on two datasets available in English and Spanish: the public MMLU benchmark and the private UNED-Access 2024 dataset. Results show that all models experience remarkable accuracy drops under our proposed variation, with an average loss of 57\% on MMLU and 50\% on UNED-Access 2024, ranging from 10\% to 93\% across models. Notably, the most accurate model in our experimentation (OpenAI-o3-mini) is not the most robust (DeepSeek-R1-70B), suggesting that the best models in standard evaluations may not be the ones with better reasoning capabilities. Also, we see larger accuracy drops in public (vs private) datasets and questions posed in their original language (vs a manual translation), which are signs of contamination and also point to a relevant role of recall/memorization in current LLMs' answers. 

\end{abstract}

\section{Introduction}

Large Language Models (LLMs) currently display remarkable performance across diverse natural language tasks, and also perform competitively with respect to humans in general knowledge benchmarks. However, a fundamental question remains: to what extent these models truly understand and reason, versus merely recalling patterns from previously seen data? This is particularly relevant in benchmarks using multiple-choice questions, which is one of the most popular methods to evaluate LLMs. While models like OpenAI’s \citep{openai2024gpt4o, openai2024o1} claim to achieve state-of-the-art performance in \textit{reasoning-heavy tasks} (such as GPQA diamond \citep{rein_gpqa_2023}), doubts persist that their success may still rely more on memorization than on the flexible reasoning that characterizes general intelligence; and reliable reasoning is crucial for many applications that require logical inference.

To examine the robustness of reasoning in LLMs, recent studies employ multi-prompt evaluations, introducing minor modifications to questions or altering mathematical problem variables. These approaches challenge models with structurally similar but unseen tasks, but are often confined to narrow domains like mathematics (see \citep{srivastava_functional_2024, mirzadeh2024gsmsymbolicunderstandinglimitationsmathematical, huang2025mathperturbbenchmarkingllmsmath}) or depend on manually curated variations, making them costly and less scalable \citep{wang_adversarial_2022}. \\



Our main goal is to investigate to what extent LLMs respond to general multiple-choice questions by searching in the (compressed) space of previously seen content, or by truly acquiring knowledge from texts and understanding the questions being posed. This leads us to work on three related research questions: \textbf{RQ1 [Reasoning vs. Memorization]}: How do models react when the questions are reformulated in a way that requires understanding and reasoning rather than recall/memorization?; \textbf{RQ2 [Contamination and translation biases]}: To what extent does prior exposure (dataset contamination) affect models' ability to reason rather than retrieve memorized answers? Additionally, how does translation impact robustness, given that translated questions are less likely to appear verbatim in training data?; and \textbf{RQ3 [Robustness predictors]}: Is the performance drop explainable just in terms of the size of the model and its (reference) effectiveness? Or there is more than scaling laws when it comes to reasoning abilities?

The main contributions of our research are: (i) we propose a simple, fully automatic method to rewrite multiple-choice questions from any domain, which ensures that they cannot be answered correctly without a genuine understanding of the subject, because the correct answer cannot be retrieved from the space of previously seen texts; (ii) we show that the performance of all models drops significantly (the average loss is above 50\%), but the drop differs widely across models; (iii) we provide additional evidence that models at least partially rely on search mechanisms for their answers, because their performance drops less with private, contamination-free datasets and with translated versions of the questions, for which recall-based answers are harder to obtain; (iv) we show that the models that perform best on the original questions are not necessarily the ones that show a smaller performance drop, indicating that reasoning capabilities and standard accuracy are not perfectly correlated. For instance, Claude-3.5-Sonnet is one of the top performers with original questions but drops up to 53\% with the reformulated questions, while DeepSeek-R1-70B is worse on the original questions but much better than Claude on the rewritten questions, which suggests that its reasoning capabilities are much higher, and (v) contrary to scaling laws that relate model size and performance, the model with the smallest drop in our experimentation is a medium-sized one (DeepSeek-R1-70B), and, in general, the models with lower drop are the most modern, particularly those integrating advanced reasoning-specific optimizations, such as o3-mini, GPT-4o and DeepSeek.


\section{Related work}

Recent advances in LLMs have raised fundamental questions about their ability to perform genuine reasoning. While some studies suggest that reasoning capabilities emerge naturally as models scale, others argue that LLMs primarily rely on memorization and statistical correlations. This section summarizes existing work on the evaluation of the reasoning capabilities of LLMs, covering general reasoning capabilities, dataset contamination issues, standard benchmarking methodologies, robustness assessments, and content variation techniques.

\subsection{LLMs and \textit{emergent reasoning capabilities}}

Reasoning is a cornerstone of general intelligence and a key criterion in LLM evaluation. While models such as GPT-4 and Claude-3 exhibit \textit{emergent capabilities} —behaviours that appear as models scale and seem to mimic reasoning—, the nature of these abilities remains debated. Many argue that LLMs mostly rely on memorized patterns and statistical associations rather than true logical inference, particularly on familiar tasks. This limitation affects their performance on out-of-distribution tasks, which are the ones that demand genuine reasoning.

\citet{smeaton2024understandingfoundationmodels1924} reviews the evolution of so-called foundation models and suggests that reasoning abilities arise not merely from increased model size, but from novel training techniques that lead to learning phenomena like grokking. This highlights two major challenges: interpreting the inner mechanisms of LLMs, and designing better evaluation methods to assess their reasoning capabilities.

\subsection{Data contamination and out-of-distribution generalization}

Data contamination complicates reasoning assessments, as distinguishing genuine reasoning requires evaluating models on unseen data —a challenge known as the out-of-distribution (OOD) generalization problem \citep{yang2023gluexevaluatingnaturallanguage}—. \citet{razeghi-etal-2022-impact} argue that evaluations ignoring pretraining data exposure are difficult to interpret, requiring a reconsideration of current benchmarking practices.

Contamination detection methods include checking dataset release dates, conducting web searches, and prompting models to reveal whether responses reflect memorized content \citep{Jiang2024InvestigatingDC, dong2024generalizationmemorizationdatacontamination, golchin2024datacontaminationquiztool, sainz-etal-2023-nlp, yang2023rethinkingbenchmarkcontaminationlanguage, samuel2024datacontaminationdetectionmodern}. However, these techniques remain limited due to ongoing model updates and indirect data leakage \citep{ahuja_mega_2023, balloccu-etal-2024-leak}. As an alternative, researchers use training data searches and controlled adversarial evaluations to mitigate contamination risks.

A popular way to mitigate contamination and better assess reasoning capabilities is to measure LLMs robustness to question variations. Generating challenging examples, often referred to as adversarial attacks, involves modifications such as using synonyms, reordering instances or introducing typos. However, these adversarial methods are often difficult to automate effectively without risking changes to the original semantic meaning \citep{wang2022adversarialgluemultitaskbenchmark}.

\subsection{Benchmarking approaches in LLMs}

General assessment of LLMs is typically conducted with question-answer datasets (often in multiple-choice format) or via \textit{LLM arenas}, where users pose their questions and compare responses from multiple LLMs \citep{chiang_chatbot_2024}, indicating their preference. Popular question-answer benchmarks include a wide range of tasks, from common sense reasoning to code generation, with exam-based assessments gaining prominence (e.g., MMLU \citep{hendrycks_measuring_2021}, GSM-8k \citep{cobbe_training_2021}, AGIEval \citep{zhong_agieval_2023}, and GPQA \citep{rein_gpqa_2023}). These benchmarks primarily focus on overall accuracy, which often cannot be directly linked to reasoning capabilities or the ability to generalize beyond training data, although benchmarks specifically designed to assess reasoning are starting to appear.

\subsection{Evaluating reasoning and robustness}

Some recent studies propose reclassifying advanced models, such as o1 (strawberry) \citep{openai2024o1}, as \textit{Large Reasoning Models} (LRMs), emphasizing the need for dedicated reasoning evaluations \citep{valmeekam2024llmscantplanlrms}. Robustness, defined as a model's ability to handle unexpected inputs, is also critical for reliable real-world applications \citep{wang2023robustnesschatgptadversarialoutofdistribution, wang2022measureimproverobustnessnlp}. \citet{embers2024} show that models struggle significantly with unseen tasks, while \citet{razeghi-etal-2022-impact} find that some GPT-based models perform disproportionately well on arithmetic problems involving frequently occurring numbers from their training data.

Beyond performance disparities, \citep{nikankin2024arithmeticalgorithmslanguagemodels} examine individual neurons in LLMs and identify circuits responsible for arithmetic operations, to conclude that LLMs neither implement robust algorithms nor rely purely on memorization, but instead apply heuristic-driven pattern matching. 

An increasing number of studies now go beyond simple accuracy metrics to assess reasoning and robustness. For instance, \citep{jiang2024peektokenbiaslarge} assess whether LLMs possess genuine reasoning abilities or primarily depend on token bias, concluding that they rely heavily on superficial patterns and struggle with logical reasoning. \citep{taghanaki2024mmluproevaluatinghigherorderreasoning} assess higher-order reasoning abilities and susceptibility to shortcut learning by introducing questions with multiple correct answers and novel metrics like the shortcut selection ratio and correct pair identification ratio, revealing significant performance disparities. Further studies highlight intrinsic limitations of models in addressing complex compositional reasoning tasks. \citet{faithandfate} find that while LLMs perform adequately on simpler problems, they struggle with systematic reasoning in more complex, multi-step tasks, often accumulating errors and failing to generalize to less common or more complex examples. As they note, ``shortcut learning via pattern-matching may yield fast correct answers when similar compositional patterns are available during training, but does not allow for robust generalization''. 

Even explicit reasoning techniques, such as Chain of Thought (CoT), may be influenced by inherent model limitations: \citet{prabhakar2024decipheringfactorsinfluencingefficacy} claim that CoT reasoning can be characterized as probabilistic, memorization-influenced noisy reasoning, indicating that LLM behaviour exhibits elements of both memorization and generalization.

\subsection{Content variation methods in reasoning evaluations}

Several studies introduce content variations to evaluate reasoning and/or detect contamination, particularly in mathematical reasoning due to its structured nature. For example, \citet{srivastava_functional_2024} generate ``functional variants'' of the MATH dataset \citep{hendrycks2021measuringmathematicalproblemsolving}, defining the \textit{reasoning gap} as the difference between \textit{static} and \textit{functional} accuracies. Similarly, \citet{mirzadeh2024gsmsymbolicunderstandinglimitationsmathematical} introduce irrelevant details into the GSM-8k dataset \citet{cobbe_training_2021}, leading to even greater accuracy drops than simple numerical variations. Likewise, \citet{hong2024evaluatingllmsmathematicalcoding} develop a semi-automatic perturbation method for mathematical and coding tasks, revealing LLMs' limited robustness to minor question modifications.

Other studies assess reasoning by analysing compositional problem dependencies. For instance, \citet{hosseini2024llmreasonerscreatedequal} examine models' performance on compositional math word problems, where solving the second problem depends on correctly answering the first, finding significant reasoning gaps, particularly in smaller or math-specialized models. Similarly, \citet{zhu_promptbench_2023} find that typos disrupt math problem-solving accuracy, while synonym changes affect sentiment analysis performance.

Beyond mathematical reasoning, some studies explore counterfactual task variants in different domains as a means to test generalization. These approaches create minimally modified yet challenging tasks, requiring the same abstract reasoning ability but with a lower likelihood of appearing frequently in an LLM’s training data. For example, \citet{wu2024reasoningrecitingexploringcapabilities} evaluate generalization by generating counterfactual variants of reasoning tasks in domains like coding and chess. Likewise, \citet{lewis2024usingcounterfactualtasksevaluate, lewis2024evaluatingrobustnessanalogicalreasoning} assess analogical reasoning abilities, and \citet{yan2024largelanguagemodelsunderstand} focus on logical reasoning. In a similar fashion to the work presented here, \citet{nezhurina_alice_2024} focus on finding one simple common sense reasoning task that can break the models, and find that even slight variations of the problem cause strong fluctuations, also expressing a strong overconfidence in the wrong solutions.\\

These studies highlight substantial performance fluctuations based on problem formulation, challenging the reliability of single-point accuracy benchmarks. In contrast to prior work, which primarily focuses on mathematical reasoning or relies on superficial prompt variations, our method introduces a universally applicable challenge that forces exhaustive answer verification. This makes it a more general and rigorous test of reasoning across disciplines.

\section{None of the others (\noto) variation}

The variation we propose on multiple-choice questions (hereafter abbreviated as \noto) is simply to replace the correct answer with "None of the other answers". This, in turn, becomes the right answer, because all other answers are incorrect by design. With this replacement, the correct answer is no longer connected terminologically or conceptually with the question, and therefore pure memorization is not enough to guess the answer. The system needs to discard all other options to conclude that "none of the others" is the correct choice. In addition, "none of the others" is frequently used in multiple-choice questions, but it tends not to be the correct answer. Therefore, a pure association memory would advise the systems not to choose "none of the others", which may lead to average results even below random choice. 

\section{Experimental setup}
\begin{figure*}[ht!]
    \centering
    \begin{minipage}{0.49\textwidth}
        \includegraphics[width=\linewidth]{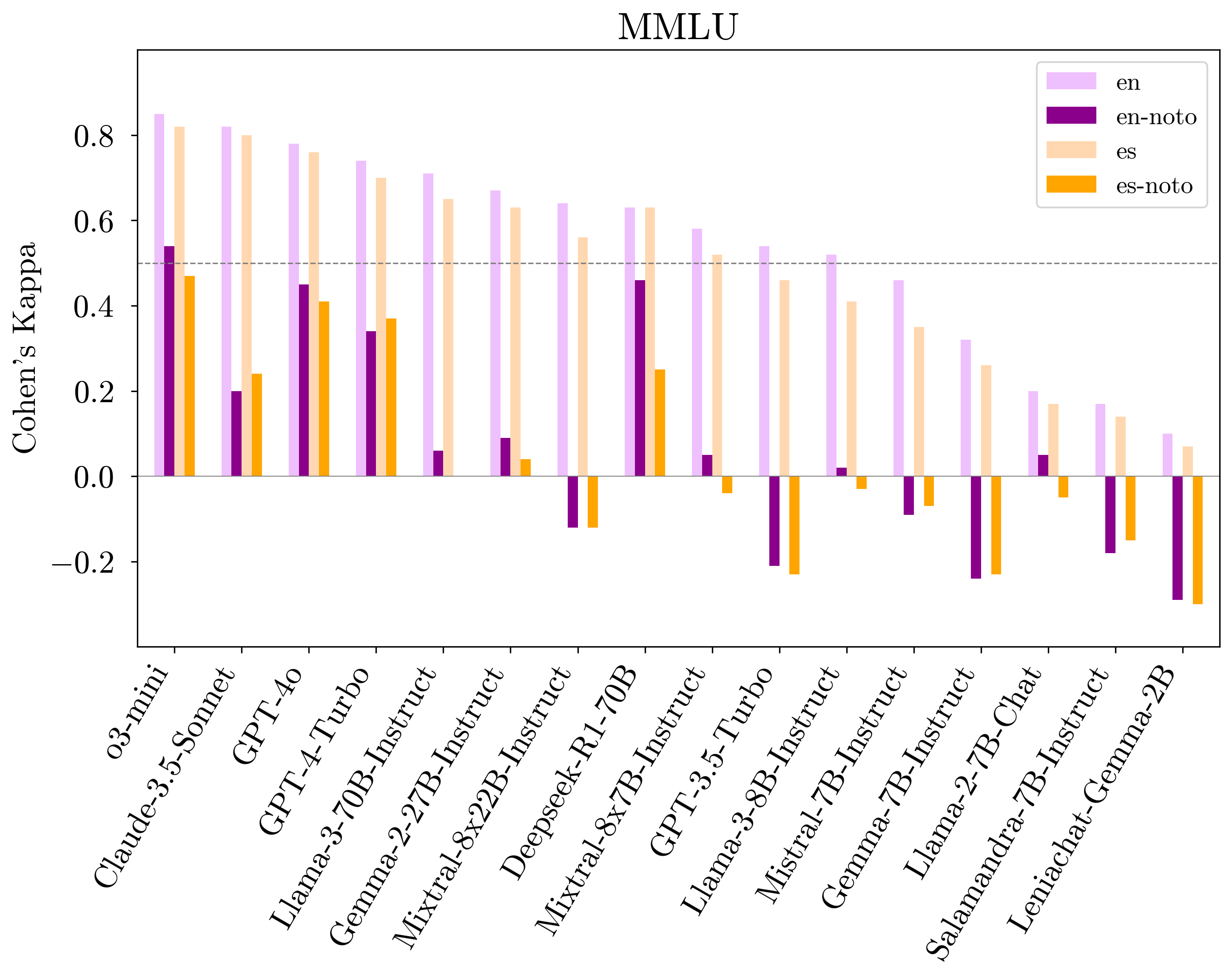}
    \end{minipage}
    \hspace{0.01\textwidth}
    \begin{minipage}{0.49\textwidth}
        \includegraphics[width=\linewidth]{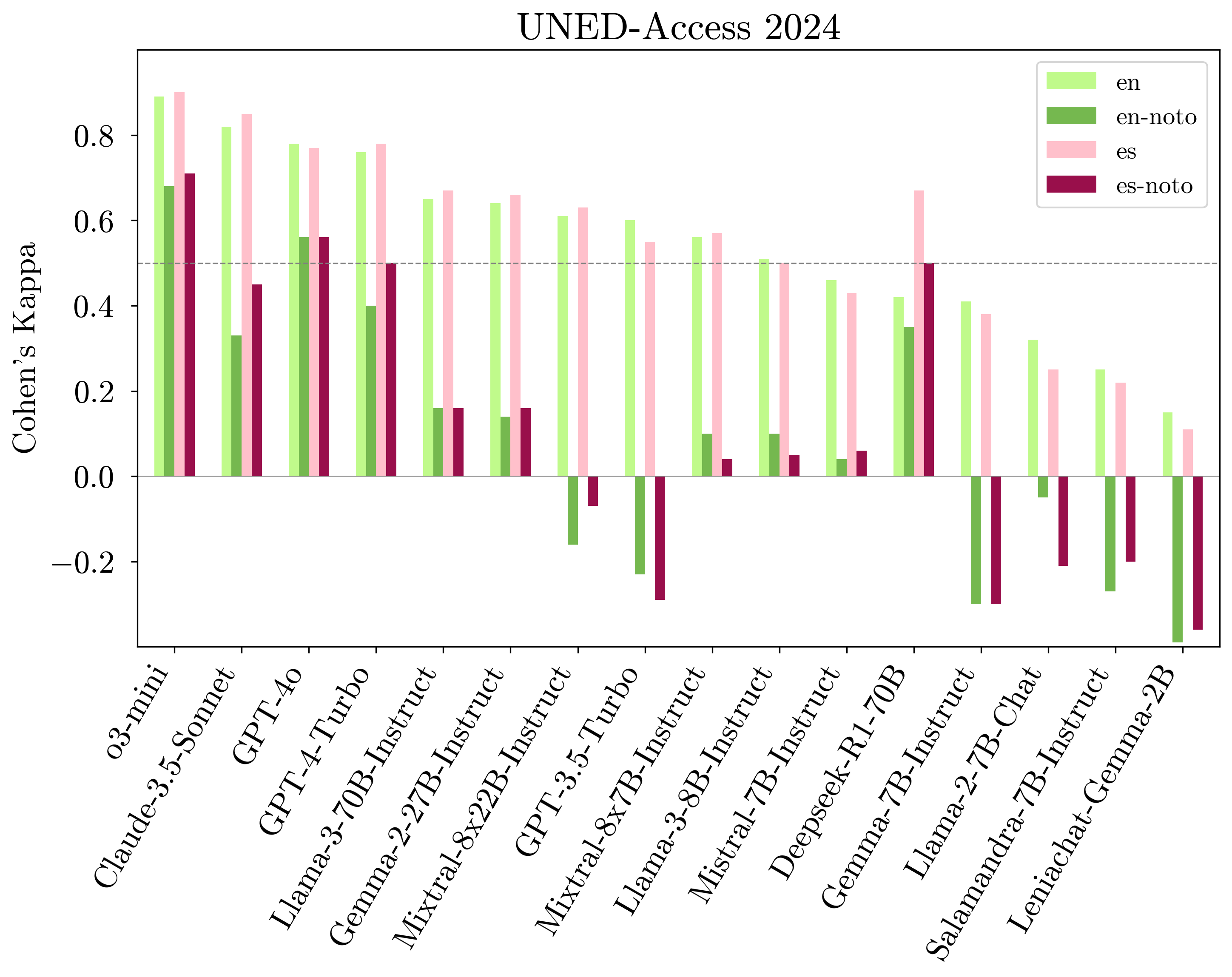}
    \end{minipage}
    \caption{Performance on MMLU and UNED-Access 2024 (original questions and \textit{none of the others} variation). Results per model and language are averaged across all subjects and expressed in terms of \textbf{Cohen's Kappa}.}
    \label{kappa}
\end{figure*}

This section provides an overview of our experimental setup, with additional details in Appendix \ref{appendix-exps} for full reproducibility. We describe the datasets, models and hyperparameters, prompting strategy, and evaluation metrics for assessing performance and robustness.

\subsection{Datasets}

We have experimented with two bilingual datasets. One is the \textbf{MMLU} dataset \citep{hendrycks_measuring_2021}, with questions in English covering 57 tasks ranging from high school to professional and graduate levels, and its professional manual translation into Spanish \citep{mmmlu}. After filtering out questions potentially incompatible with the ``none of the others'' substitution, this dataset comprises 13,346 questions.

The second is \textbf{UNED-Access 2024}, with 1,003 questions in Spanish on 11 university-entry-level subjects, and professional manual translations into English. Unlike MMLU, this dataset has never been made public and therefore the effects of contamination should be minimal. After preprocessing, 950 compatible questions remained for the \noto scenario.


\subsection{Models and hyperparameters}

Experiments were conducted using 15 generative models, including five proprietary models and ten open-source models, all trained for instruction following. Proprietary models were accessed via API, while the open-source models were obtained from Hugging Face or deployed via the Ollama\footnote{\url{https://ollama.com/}} library.

The temperature was set to 0 for all models to ensure deterministic outputs (except for o3-mini, which did not allow temperature adjustment) to minimize creativity and focus on factual or reasoning-based answers. Each question was provided one at a time using a fixed prompt structure, which included a system prompt, user prompt, and assistant prompt. The prompt specified the subject of the question and was always in the same language as the dataset (English or Spanish) as is done in other evaluations \citep{zhang_m3exam_2023, openai2024gpt4technicalreport}.  

\subsection{Prompting strategy}

LLM evaluations often mix prompting strategies, ranging from zero-shot to few-shot (with varying numbers of examples) and Chain-of-Thought configurations. We adopt a uniform zero-shot setting, as it closely resembles real-world interactions with LLMs while ensuring a simpler and more replicable experimental setup, and may even lead to better results for the most recent models \citep{deepseekai2025deepseekr1incentivizingreasoningcapability}.

\subsection{Evaluation metrics}

Most LLM evaluations rely on \textbf{Accuracy}, calculated as the proportion of correct answers ($C$) over the total responses ($N$). However, Accuracy alone does not account for variations in the number of answer choices or chance-level performance. Since different subjects in our datasets have varying numbers of answer choices, we complement Accuracy with \textbf{Cohen’s Kappa coefficient}, which accounts for chance-level performance and enables fairer comparisons across subjects:

\vspace{-0.25cm}
\begin{equation*}
\resizebox{\columnwidth}{!}{$
\text{Kappa} = \frac{\text{observed accuracy} - \text{expected accuracy}}{1 - \text{expected accuracy}}
= \frac{\frac{\text{C}}{\text{N}} - \frac{1}{\text{M}}}{1 - \frac{1}{\text{M}}}
$}
\end{equation*}

where $M$ is the number of possible choices and the \textit{expected accuracy} corresponds to random guessing: 1/3 for three-choice questions and 1/4 for four-choice questions. Cohen’s Kappa normalizes correctness so that random answers yield a Kappa of zero, enabling fair comparisons across subjects. Kappa values range from -1/2 to 1, and negative values indicate performance worse than random chance. The final result for each model and language is the arithmetic mean of Kappa values across all subjects, giving equal weight to each subject to account for dataset imbalances.

To measure performance variation between original and modified questions, which we refer to as \textit{drop}, we report the percentage decrement of Accuracy\footnote{Cohen's Kappa coefficient is not in a ratio scale (the origin is not zero) and therefore percentages cannot be computed directly.}  





\section{Results}


Each question from MMLU and UNED-Access 2024 is evaluated under four conditions: original formulation in English and Spanish, and a modified version where the correct answer is replaced with ``None of the other answers'' in both languages. This setup enables direct comparison between standard multiple-choice performance and the reasoning challenge introduced by the exclusion option.

Figure \ref{kappa} provides a visual representation of performance in terms of Cohen's Kappa (detailed numerical results are available in Appendix \ref{app-results}), while Table \ref{tabla_accuracy_drops} presents accuracy scores and performance drop between the original questions and their \noto variations. 

\begin{table*}[ht!]
    \centering
    \resizebox{\textwidth}{!}{%
    \begin{tabular}{lccc|ccc|ccc|ccc}
        \toprule
        & \multicolumn{3}{c|}{MMLU (English)} & \multicolumn{3}{c|}{MMLU (Spanish)} & \multicolumn{3}{c|}{UNED (English)} & \multicolumn{3}{c}{UNED (Spanish)} \\
        \cmidrule(lr){2-4} \cmidrule(lr){5-7} \cmidrule(lr){8-10} \cmidrule(lr){11-13}
         & base & \noto & drop \% & base & \noto & drop \% & base & \noto & drop \% & base & \noto & drop \% \\
        \midrule
        DeepSeek-R1-70B & 0.72 & 0.60 & \underline{16.67} & 0.72 & 0.44 & 38.89 & 0.60 & 0.54 & \underline{10.0} & 0.77 & 0.65 & 15.58 \\
        OpenAI-o3-mini & \textbf{0.89} & \textbf{0.65} & 26.97 & \textbf{0.86} & \textbf{0.6} & \underline{30.23} & \textbf{0.92} & \textbf{0.77} & 16.30 & \textbf{0.93} & \textbf{0.79} & \underline{15.05} \\
        Llama-2-7B-Chat & 0.40 & 0.29 & 27.50 & 0.38 & 0.21 & 44.74 & 0.52 & 0.26 & 50.00 & 0.48 & 0.15 & 68.75 \\
        GPT-4o & 0.84 & 0.58 & 30.95 & 0.82 & 0.56 & 31.71 & 0.85 & 0.69 & 18.82 & 0.84 & 0.69 & 17.86 \\
        GPT-4-Turbo & 0.81 & 0.51 & 37.04 & 0.78 & 0.53 & 32.05 & 0.84 & 0.58 & 30.95 & 0.85 & 0.65 & 23.53 \\
        Claude-3.5-Sonnet & 0.86 & 0.40 & 53.49 & 0.85 & 0.43 & 49.41 & 0.87 & 0.53 & 39.08 & 0.89 & 0.62 & 30.34 \\
        Gemma-2-27B-Instruct & 0.75 & 0.32 & 57.33 & 0.72 & 0.28 & 61.11 & 0.75 & 0.40 & 46.67 & 0.76 & 0.41 & 46.05 \\
        Mixtral-8x7B-Instruct & 0.68 & 0.29 & 57.35 & 0.64 & 0.22 & 65.63 & 0.69 & 0.37 & 46.38 & 0.70 & 0.33 & 52.86 \\
        Llama-3-8B-Instruct & 0.64 & 0.26 & 59.38 & 0.56 & 0.23 & 58.93 & 0.66 & 0.37 & 43.94 & 0.65 & 0.33 & 49.23 \\
        Llama-3-70B-Instruct & 0.78 & 0.30 & 61.54 & 0.73 & 0.25 & 65.75 & 0.75 & 0.41 & 45.33 & 0.77 & 0.41 & 46.75 \\
        Mistral-7B-Instruct & 0.59 & 0.18 & 69.49 & 0.51 & 0.20 & 60.78 & 0.62 & 0.33 & 46.77 & 0.60 & 0.34 & 43.33 \\
        Salamandra-7B-Instruct & 0.38 & 0.11 & 71.05 & 0.36 & 0.14 & 61.11 & 0.47 & 0.11 & 76.60 & 0.46 & 0.16 & 65.22 \\
        Mixtral-8x22B-Instruct & 0.73 & 0.16 & 78.08 & 0.67 & 0.16 & 76.12 & 0.73 & 0.19 & 73.97 & 0.74 & 0.25 & 66.22 \\
        Gemma-7B-Instruct & 0.49 & 0.07 & 85.71 & 0.44 & 0.08 & 81.82 & 0.59 & 0.09 & 84.75 & 0.56 & 0.09 & 83.93 \\
        GPT-3.5-Turbo & 0.66 & 0.09 & 86.36 & 0.60 & 0.08 & 86.67 & 0.72 & 0.14 & 80.56 & 0.68 & 0.10 & 85.29 \\
        Leniachat-Gemma-2B & 0.32 & 0.03 & 90.63 & 0.30 & 0.03 & 90.00 & 0.40 & 0.03 & 92.50 & 0.37 & 0.05 & 86.49 \\
        \bottomrule
    \end{tabular}
    }
    \caption{\textbf{Accuracy} results on the original and \textit{none of the others} configurations, and percentage decrease between scenarios. Systems are sorted by drop in English MMLU, smaller to largest.}
    \label{tabla_accuracy_drops}
\end{table*}

\subsection{RQ1: Performance vs. Reasoning}

To evaluate the impact of the exclusion option on model performance, we analyse both effectiveness (Cohen's Kappa and Accuracy) and robustness across datasets in English and Spanish. 

\textbf{Performance with \noto:} Figure \ref{kappa} depicts results in terms of Cohen's Kappa (where random guessing always gets zero regardless of the number of choices)\footnote{Note that the slightly lower MMLU results compared to previous studies are primarily due to our use of Cohen's Kappa. Additional differences may stem from our zero-shot setup (versus few-shot in other works), prompt formulation, or the quantization of Ollama models.}. All models exhibit a substantial drop in performance under the \noto variation. In multiple cases, models are even worse than random answers, which suggests that they rely almost purely on memorization, and they probably learnt in the pre-training phase that "None of the others" is statistically less likely than any other option. With the \noto variation, only o3-mini (the top-performing model overall) exceeds the 0.5 passing threshold in MMLU, for one language (English). Note that, with the use of appropriate question variations, the MMLU dataset is far from being saturated. For UNED-Access 2024, two models pass in English (o3-mini and GPT-4o) and four in Spanish (o3-mini, GPT-4o, GPT-4-Turbo, and DeepSeek). 



\textbf{Performance drop:} The accuracy drop (Table \ref{tabla_accuracy_drops}) varies drastically across models (from 10\% to 92.5\%) in all four datasets, highlighting substantial differences in robustness. 
Some mid-sized models such as Mixtral-8x22B and GPT-3.5-Turbo suffer particularly steep drops comparable to much smaller models, and scores well below random chance in the \noto setting. The same applies to somewhat more modern models, such as Llama-3-70B and Gemma-2-27B, which fall drastically to near random-chance performance. Among the top performing models, Claude-3.5-Sonnet experiences the most remarkable drop: despite achieving strong performance in the original setting, its \noto accuracy falls well below that of its peers (o3-mini, DeepSeek-R1-70B, GPT-4-Turbo, and GPT-4o).

DeepSeek's R1 case is particularly surprising: although the 70B model ranks well below the top performers on the original dataset, it exhibits the smallest accuracy drop in both English datasets, and also the lowest drop overall (only 10\% in UNED-Access 2024 in English and 16.67\% in English MMLU). This suggests that while DeepSeek-R1-70b is smaller and with less memory, it has stronger reasoning abilities. 

Overall, these results reveal significant differences in how models handle scenarios that demand refined reasoning. DeepSeek-R1-70B and OpenAI-o3-mini has the smallest relative drops, which indicates a stronger, albeit imperfect, ability to validate answer options rather than rely solely on memorization. In contrast, Claude-3.5-Sonnet, despite being a high-performing model in standard conditions, suffers one of the largest drops (53.49\% in English MMLU). The most affected models, such as GPT-3.5-Turbo, experience extreme accuracy degradation (over 85\% drop), which points to an almost exclusive reliance on approximate matching heuristics.




\subsection{RQ2: Contamination and translation biases}
\begin{figure*}[ht!]
    \centering
    \includegraphics[width=\linewidth]{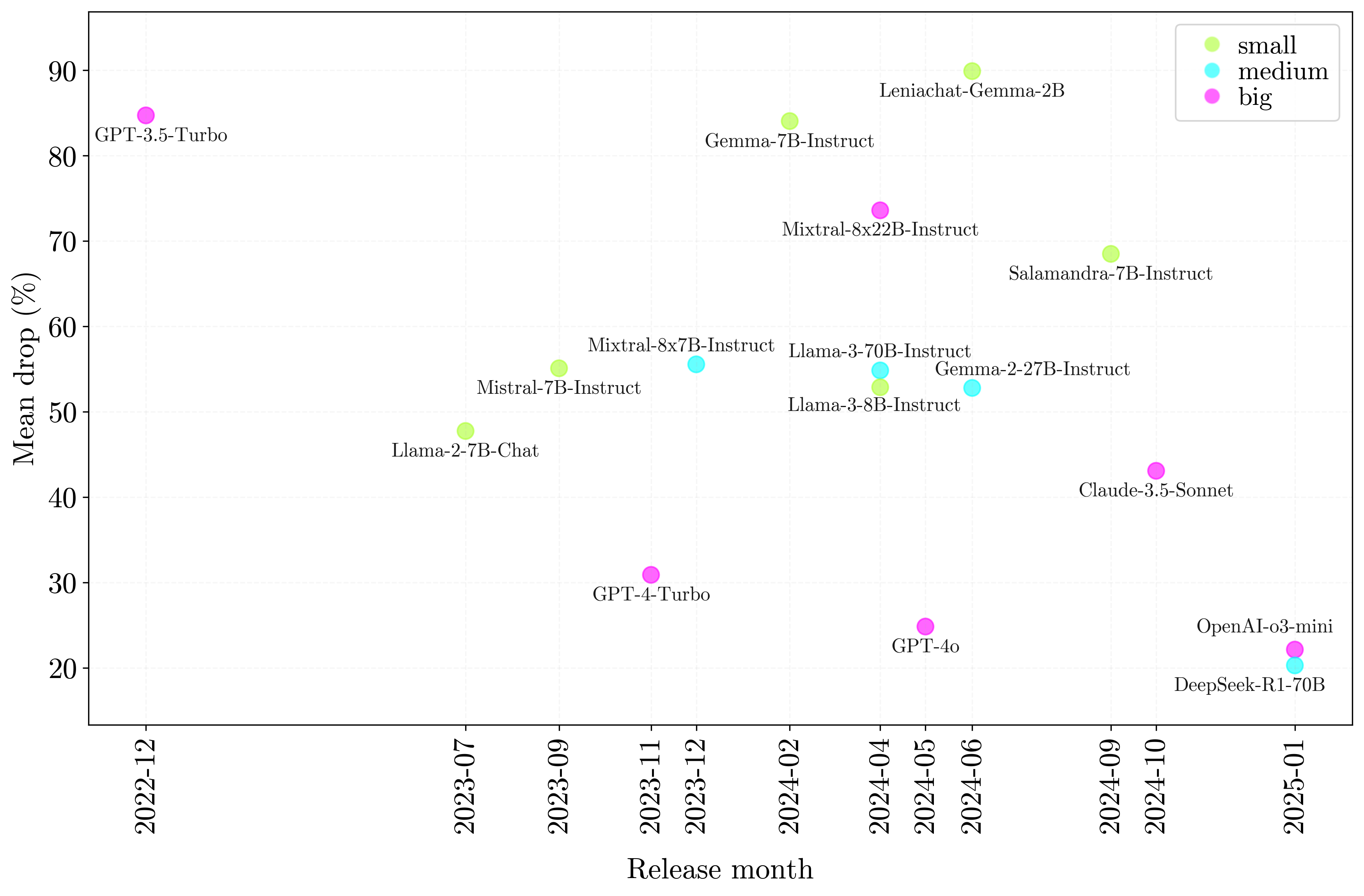}
    \caption{\centering Mean drop across release dates and model sizes.}
    \label{drop_month}
\end{figure*}

To investigate whether the accuracy drop is due to reasoning limitations or reliance on memorized patterns, we compare results from two angles: (1) the effect of dataset contamination, contrasting the public MMLU dataset with the private UNED-Access 2024 dataset, and (2) the effect of translations, contrasting models' performance in the original language versus manually translated versions. These aspects are closely related, as they both influence the extent to which models rely on prior exposure rather than true reasoning. If memorization plays a dominant role, we expect larger drops in public datasets and in original-language versions, as these are more likely to have been seen during pretraining, and for them approximate search may be more effective.

\textbf{Contamination effects:} The mean drop is higher in MMLU (56.85\%) than in UNED-Access 2024 (49.78\%), consistent with the expectation that MMLU, as a public dataset, is more likely to have been seen during pretraining and leads models to fail more when they are prevented from using that memorisation. In fact, the lowest absolute drop (10\% from DeepSeek) is observed on the least likely contaminated dataset, the English UNED-Access 2024 (which is both private and translated from the original questions).

\textbf{Translation effects:} Within MMLU, the average drop is slightly greater in Spanish (58.43\%) than in English (56.85\%), whereas in UNED-Access 2024, the pattern reverses (50.16\% in English vs. 49.78\% in Spanish). With the original questions, models perform better in each dataset’s original language: all models (15/15) achieve higher accuracy in English for MMLU, while in UNED-Access 2024, 8 models perform better in Spanish. This trend still holds in the \noto scenario: in MMLU, 8 models perform better in English, and in UNED-Access 2024, 9 models now perform better in Spanish. When considering passing thresholds, more models pass in English for MMLU (11 vs. 9), and in Spanish for UNED-Access 2024 (11 vs. 10). This pattern holds in \noto: 1 vs. 0 in MMLU and 4 vs. 2 in UNED-Access 2024. 


These are signs of contamination, since, in other words, (i) models fall more in the public dataset, which is likely more contaminated and (ii) models fall more in the original versions than in the translated versions, with which models are probably less familiar (the Spanish MMLU is newer and less likely to be contaminated, and even if UNED-Access 2024 is private, it is less likely that models have seen the English questions since they are manual translations and have never been released). 

Overall, these findings confirm that the \noto substitution exposes reliance on memorized content, and lead us to the conclusion that models experiencing the highest drops are those most likely answering with their memorization skills, rather than with true reasoning. Results with the \noto configuration are a better indication of models' true capabilities, show that with a little twikering the datasets are far from being saturated, and reveal comparative differences between the reasoning capabilities of models that are hidden in the evaluation with the original questions. In particular, we have seen that the performance difference between the most recent \textit{reasoning} models (Deepkseek, o3-mini) and other state-of-the-art ones such as Claude-3.5 is much wider than can be measured with the original questions.


 



\subsection{RQ3: Robustness predictors}

\begin{table}[ht!]
\centering
\footnotesize 
\resizebox{0.9\columnwidth}{!}{
\begin{tabular}{|l|c|c|}
\hline
\textbf{} & \textbf{correlation} & \textbf{$p$-value} \\ \hline
MMLU (English) & -0.47 & 0.0667 \\ \hline
MMLU (Spanish) & -0.59 & 0.0165 \\ \hline
UNED (English) & -0.60 & 0.0130 \\ \hline
UNED (Spanish) & -0.78 & 0.0003 \\ \hline
\end{tabular}%
}
\caption{Pearson's correlation between accuracy results on the base configuration and the drop.}
\label{tab:pearson_correlation}
\end{table}

Here we try to answer the question: are there model features that can predict greater robustness? Table \ref{tab:pearson_correlation} shows Pearson's correlation between results on the base configuration and the drop. 
The table indicates that, while there is a substantial inverse correlation between effectiveness on the original questions and performance drop with the \noto variation, it is not entirely reliable as a predictor; and in one of the datasets (English MMLU) is not even statistically significant. As we saw before, what we think are the best models using the original questions may not be the ones reasoning better. 

Figure \ref{drop_month} shows the the mean drop in performance across models, sorted by release date and classified into three groups according to their size: small (less than 10B parameters), medium (10-100B) and large (over 100B). Note that the drop does not correlate well with model size, as there are large models with large drops (GPT-3.5-Turbo, Mixtral-8x22B and Claude-3.5-Sonnet), and the smallest average drop is for a medium-sized model (DeepSeek-R1-70B): size alone is insufficient to ensure robust reasoning. There is a noticeable trend, though, where newer models tend to exhibit smaller drops, with some exceptions. The oldest model, GPT-3.5-Turbo, is a mid performer with the original datasets, but stands out as one of the worst in terms of performance drop. In the period since ChatGPT's debut, the generalisation capabilities of models seems to have improved widely and consistently, and this improvement does not necessarily come with increased model sizes. Finally, the newest proprietary models and DeepSeek-R1 are the ones that show smaller performance drops; this suggests that robustness in reasoning is influenced more by advanced model architectures and training strategies rather than sheer model size.

\section{Conclusions}
Our results show that the proposed \noto variation poses a major challenge for LLMs, and provides a useful signal to distinguish answers based on recall/memorization from genuine knowledge and reasoning. While many models perform well when retrieving memorized information, their performance plummets when the correct answer is disconnected from memory associations and they are required to verify and reject each candidate answer. The \noto variation consistently reveals reasoning gaps, exposing limitations that remain hidden in standard multiple-choice settings (RQ1). Dataset contamination further complicates the evaluation of reasoning: while prior exposure may artificially inflate accuracy in base scenarios, its impact diminishes in \noto, where models cannot rely on memorized answers. Similarly, models perform better on original (and likely more contaminated) datasets, while translated versions mitigate this effect, reinforcing the role of memorization in standard benchmarks (RQ2).

Unlike accuracy, which scaling laws correlate with model size \citep{kaplan2020scalinglawsneurallanguage}, we have seen that robustness is not strictly correlated with model size. High-performing models such as Claude-3.5 suffer severe drops, and some mid-sized models (e.g., GPT-3.5-Turbo, Mixtral) degrade to below-random performance. The most robust model in our experimentation, DeepSeek-R1-70B, is mid-sized, suggesting that architectural advancements and training strategies, rather than sheer scale, play a greater role in reasoning robustness (RQ3). Remarkably, the two so-called \textit{reasoning models} in our sample (o3-mini and DeepSeek-R1) are indeed the ones that better resist the \noto variation. 

In short, our experimentation is a direct confirmation that LLMs remain far from true reasoning, but also that progress is being made towards that goal. Our findings emphasize the need for models that can reliably handle question reformulations without relying on surface-level heuristics, and show that classic datasets that appear to be saturated, such as MMLU, may still be useful for LLM evaluation under appropriate transformations. 


\section*{Limitations}
Our evaluation relies on multiple-choice questions, which cannot give a comprehensive evaluation of reasoning abilities. Our results are meant to be complementary with other research directions in evaluation of LLMs. Additionally, our findings are based on MMLU and UNED-Access 2024; while they give an interesting combination of private \& public data, and bilingualism in two directions, different knowledge domains or difficulty patterns may lead to different conclusions. Expanding evaluation to other benchmarks, including those focused on LLM performance in real-world tasks, will provide a broader perspective of LLMs' performance.


Although humans also use pattern-matching for reasoning \citep{dasgupta2024languagemodelshumanlikecontent}, their cognitive processes differ fundamentally from those of LLMs. Future work should compare human performance on exclusion-based evaluations to better understand these differences.

Finally, our experiments follow a zero-shot setting to ensure consistency and replicability, but alternative prompting techniques—such as Chain-of-Thought or few-shot reasoning—could may yield different outcomes. Further studies could explore whether structured reasoning prompts help mitigate the observed performance drop.



\section*{Acknowledgments}

This work has been funded by the European Union - NextGenerationEU through the `Recovery, Transformation and Resilience Plan', by the Ministry of Economic Affairs and Digital Transformation and by UNED-Access 2024. However, the views and opinions expressed are solely those of the author(s) and do not necessarily reflect those of the European Union or the European Commission. Neither the European Union nor the European Commission can be held responsible for them.

\bibliography{references}

\appendix

\section{Experimental setup}
\label{appendix-exps}

\subsection{Datasets}

The replacement of the correct answer with ``None of the other answers'' requires the question to be compatible, meaning that the answer choices cannot already include options such as ``None of the Above'', ``All of the above'', or similar formulations that would interfere with the intended substitution. To ensure this, we ran a script using regular expressions to automatically detect and filter out questions with incompatible answer choices.

For MMLU, this filtering resulted in 13,470 compatible questions in English and 13,449 in Spanish. To maintain consistency across versions, we used the 13,346 questions that appeared in both languages for our experiments.

In \textit{UNED-Access 2024}, the same methodology was applied with one exception: since almost all psychology questions included ``None of the other answers'' as the fourth option, we opted to modify rather than discard them. Specifically, we removed this answer option unless it was the correct answer, in which case we discarded the question. This resulted in 950 compatible questions for the ``None of the others'' variation, while the original 1,003 questions remained unchanged for the base setting.

\subsection{Models and hyperparameters}

These were the models used for the experimentation: 

\begin{itemize}[label={},leftmargin=0cm,itemsep=0.2cm, topsep=0.1cm, parsep=0.1cm] 
    \item \textbf{Proprietary models:}
    \begin{itemize}[itemsep=0.1cm,parsep=0cm,topsep=0cm] 
        \item[\raisebox{-0.2\height}{\includegraphics[height=0.8em]{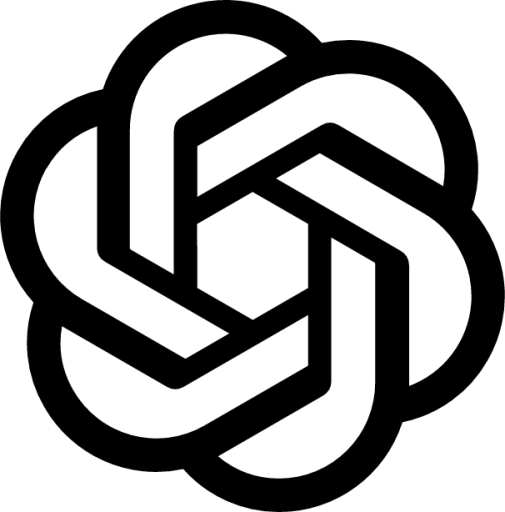}}] via \textbf{OpenAI} API: o3-mini \citep{o3-mini-2025}, GPT-4-Turbo \citep{openai2024gpt4technicalreport}, GPT-4o \citep{openai2024gpt4o}, GPT-3.5-Turbo \citep{brown2020languagemodelsfewshotlearners}.
        \item[\raisebox{-0.2\height}{\includegraphics[height=0.8em]{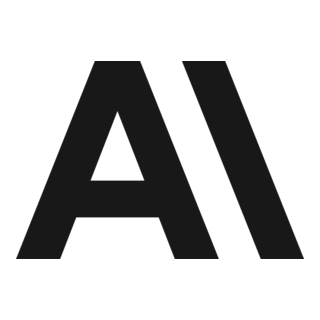}}] via \textbf{Anthropic} API: Claude-3.5-Sonnet \citep{claude_2024}.
    \end{itemize}

    \item \textbf{Open-source models:}
    \begin{itemize}[itemsep=0.1cm,parsep=0cm,topsep=0cm] 
        \item[\raisebox{-0.2\height}{\includegraphics[height=0.8em]{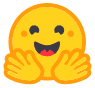}}] \textbf{Hugging Face}: Llama-2-7B \citep{touvron_llama_2023}, Llama-3-8B \citep{llama_3_2024}, Gemma-7B \citep{gemmateam2024gemmaopenmodelsbased}, Gemma-2-27B, Mistral-7B \citep{jiang_mistral_2023}, Leniachat-Gemma-2B\footnote{\url{https://huggingface.co/LenguajeNaturalAI/leniachat-gemma-2b-v0}}, Salamandra-7B\footnote{\url{https://huggingface.co/LenguajeNaturalAI/leniachat-gemma-2b-v0}}.
        \item[\raisebox{-0.2\height}{\includegraphics[height=0.8em]{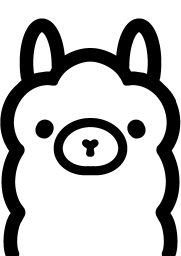}}] \textbf{Ollama}: DeepSeek-R1-70B \citep{deepseekai2025deepseekr1incentivizingreasoningcapability}, Llama-3-70B \citep{llama_3_2024}, Mixtral-8x7B and Mixtral-8x22B \citep{jiang2024mixtralexperts}, Gemma-2-27B \citep{gemmateam2024gemma2improvingopen}.
    \end{itemize}
\end{itemize}

\subsection*{Prompting strategy}
The exact prompts used were:

{\small
\renewcommand{\labelitemi}{{\textcolor[HTML]{9999ff}{\scriptsize\ding{117}}}}
\begin{itemize}
\setlength\itemsep{0.01em}

\item {\bf System prompt} \newline {\bf ES:} \texttt{Eres un sistema experto en responder preguntas de exámenes.} \newline {\bf EN:} \texttt{You are an expert system for answering exam questions.}

\item {\bf User prompt} \newline {\bf ES:} \texttt{Responde a la siguiente pregunta de la asignatura \{\}, tan solo con la letra de la respuesta correcta. Pregunta: \{\}} \newline {\bf EN:} \texttt{Answer the following question of the subject \{\} only with the letter of the correct answer. Question: \{\}}

\item {\bf Assistant prompt} \newline {\bf ES:} \texttt{Letra de la respuesta correcta:} \newline {\bf EN:} \texttt{Letter of the correct answer:}
\end{itemize}
}

For open models, the instructions were formatted according to their respective training specifications, as detailed in their model cards. Finally, responses were post-processed to extract the predicted answer, removing any additional justifications or extraneous text before evaluation.

\section{Results}
\label{app-results}

Table \ref{tablauned} presents Cohen's Kappa results for \textit{UNED-Access 2024} by subject, while Tables \ref{tablammlu1} to \ref{tablammlu4} display the MMLU results, each table corresponding to a different subject category.

\begin{table*}[ht]
\centering
\resizebox{\linewidth}{!}{%
\renewcommand{\arraystretch}{0.8}
\begin{tabular}{lcccccccccccc}

 & \textbf{\small\begin{sideways}BAM\end{sideways}} & \textbf{\small\begin{sideways}Biology\end{sideways}} & \textbf{\small\begin{sideways}Biochemistry\end{sideways}} & \textbf{\small\begin{sideways}Economics\end{sideways}} & \textbf{\small\begin{sideways}F. of Computing\end{sideways}} & \textbf{\scriptsize\begin{sideways}Spanish Language\end{sideways}} & \textbf{\small\begin{sideways}Literature\end{sideways}} & \textbf{\small\begin{sideways}Mathematics\end{sideways}} & \textbf{\scriptsize\begin{sideways}Math Applied to SS\end{sideways}} & \textbf{\small\begin{sideways}Advanced Math  
\end{sideways}} & \textbf{\small\begin{sideways}Psychology\end{sideways}} & \textbf{Average}\\
\toprule
ENGLISH-original  &&&&&&&&&&&&\\
\midrule
o3-mini & 0.79 & 0.96 & 1.00 & 0.90 & 0.94 & 0.76 & 0.65 & 0.98 & 0.94 & 1.00 & 0.85 & 0.89 \\
Claude-3.5-Sonnet & 0.83 & 0.95 & 1.00 & 0.95 & 0.94 & 0.83 & 0.84 & 0.55 & 0.66 & 0.56 & 0.88 & 0.82 \\
GPT-4o & 0.79 & 0.96 & 1.00 & 0.92 & 0.96 & 0.70 & 0.82 & 0.55 & 0.52 & 0.50 & 0.86 & 0.78 \\
GPT-4-Turbo & 0.78 & 0.97 & 1.00 & 0.92 & 0.94 & 0.67 & 0.74 & 0.55 & 0.51 & 0.50 & 0.83 & 0.76 \\
Llama-3-70B-Instruct & 0.74 & 0.90 & 1.00 & 0.82 & 0.92 & 0.33 & 0.60 & 0.34 & 0.44 & 0.19 & 0.82 & 0.65 \\
Gemma-2-27B-Instruct & 0.72 & 0.94 & 1.00 & 0.79 & 0.87 & 0.50 & 0.55 & 0.28 & 0.33 & 0.25 & 0.81 & 0.64 \\
Mixtral-8x22B-Instruct & 0.66 & 0.90 & 0.95 & 0.79 & 0.81 & 0.40 & 0.59 & 0.30 & 0.28 & 0.25 & 0.81 & 0.61 \\
GPT-3.5-Turbo & 0.67 & 0.84 & 0.95 & 0.61 & 0.89 & 0.36 & 0.56 & 0.28 & 0.17 & 0.50 & 0.73 & 0.60 \\
Mixtral-8x7B-Instruct & 0.71 & 0.81 & 0.92 & 0.61 & 0.87 & 0.32 & 0.52 & 0.22 & 0.33 & 0.13 & 0.73 & 0.56 \\
Llama-3-8B-Instruct & 0.52 & 0.77 & 0.90 & 0.61 & 0.79 & 0.38 & 0.43 & 0.20 & 0.28 & 0.13 & 0.67 & 0.51 \\
Mistral-7B-Instruct & 0.57 & 0.71 & 0.82 & 0.63 & 0.77 & 0.23 & 0.36 & 0.05 & 0.23 & -0.00 & 0.65 & 0.46 \\
Deepseek-R1-70B & 0.64 & 0.75 & 0.62 & 0.74 & 0.58 & 0.30 & 0.49 & 0.03 & -0.10 & -0.06 & 0.66 & 0.42 \\
Gemma-7B-Instruct & 0.41 & 0.67 & 0.85 & 0.56 & 0.75 & 0.18 & 0.22 & 0.20 & 0.14 & -0.06 & 0.61 & 0.41 \\
Llama-2-7B-Chat & 0.43 & 0.62 & 0.39 & 0.27 & 0.62 & 0.15 & 0.30 & 0.12 & 0.15 & -0.00 & 0.48 & 0.32 \\
Salamandra-7B-Instruct & 0.43 & 0.32 & 0.49 & 0.19 & 0.68 & 0.09 & 0.30 & 0.08 & -0.04 & -0.06 & 0.22 & 0.25 \\
Leniachat-Gemma-2B & 0.29 & 0.32 & 0.24 & 0.03 & 0.13 & 0.05 & 0.22 & 0.08 & 0.07 & -0.06 & 0.27 & 0.15 \\
\midrule
ENGLISH-noto &&&&&&&&&&&&\\
\midrule
o3-mini & 0.54 & 0.90 & 0.89 & 0.75 & 0.81 & 0.43 & 0.05 & 0.98 & 0.89 & 1.00 & 0.26 & 0.68 \\
GPT-4o & 0.53 & 0.95 & 0.89 & 0.68 & 0.77 & 0.43 & 0.12 & 0.40 & 0.31 & 0.44 & 0.68 & 0.56 \\
GPT-4-Turbo & 0.37 & 0.79 & 0.84 & 0.51 & 0.77 & 0.28 & 0.00 & 0.32 & 0.20 & -0.06 & 0.40 & 0.40 \\
Deepseek-R1-70B & 0.53 & 0.69 & 0.43 & 0.61 & 0.30 & 0.49 & 0.36 & 0.22 & 0.03 & 0.31 & -0.11 & 0.35 \\
Claude-3.5-Sonnet & 0.11 & 0.75 & 0.86 & 0.58 & 0.66 & 0.38 & 0.11 & 0.10 & 0.04 & -0.25 & 0.32 & 0.33 \\
Llama-3-70B-Instruct & 0.07 & 0.40 & 0.62 & 0.40 & 0.37 & -0.04 & -0.23 & 0.03 & 0.04 & -0.12 & 0.18 & 0.16 \\
Gemma-2-27B-Instruct & 0.13 & 0.50 & 0.67 & 0.40 & 0.53 & -0.01 & -0.04 & -0.07 & -0.23 & -0.38 & 0.03 & 0.14 \\
Llama-3-8B-Instruct & -0.04 & 0.31 & 0.26 & -0.02 & 0.28 & 0.08 & -0.26 & 0.24 & -0.15 & 0.25 & 0.19 & 0.10 \\
Mixtral-8x7B-Instruct & 0.01 & 0.28 & 0.43 & 0.23 & 0.26 & 0.06 & -0.23 & 0.05 & -0.16 & -0.00 & 0.19 & 0.10 \\
Mistral-7B-Instruct & 0.13 & 0.06 & 0.18 & 0.19 & 0.28 & -0.01 & -0.17 & -0.15 & -0.12 & -0.06 & 0.12 & 0.04 \\
Llama-2-7B-Chat & -0.03 & 0.01 & -0.15 & -0.19 & -0.02 & -0.04 & 0.02 & -0.15 & 0.01 & -0.06 & 0.09 & -0.05 \\
Mixtral-8x22B-Instruct & -0.23 & -0.21 & 0.05 & 0.05 & -0.02 & -0.04 & -0.25 & -0.32 & -0.32 & -0.31 & -0.14 & -0.16 \\
GPT-3.5-Turbo & -0.35 & -0.27 & -0.12 & -0.16 & -0.10 & -0.08 & -0.30 & -0.27 & -0.37 & -0.31 & -0.16 & -0.23 \\
Salamandra-7B-Instruct & -0.39 & -0.37 & -0.17 & -0.12 & -0.08 & -0.29 & -0.16 & -0.27 & -0.40 & -0.50 & -0.23 & -0.27 \\
Gemma-7B-Instruct & -0.29 & -0.34 & -0.09 & -0.30 & -0.10 & -0.21 & -0.30 & -0.46 & -0.42 & -0.50 & -0.26 & -0.30 \\
Leniachat-Gemma-2B & -0.39 & -0.45 & -0.45 & -0.30 & -0.29 & -0.30 & -0.33 & -0.44 & -0.48 & -0.50 & -0.32 & -0.39 \\
\midrule
SPANISH-original  &&&&&&&&&&&&\\
\midrule
o3-mini & 0.90 & 0.96 & 1.00 & 0.92 & 0.94 & 0.82 & 0.56 & 1.00 & 0.97 & 1.00 & 0.82 & 0.90 \\
Claude-3.5-Sonnet & 0.93 & 0.96 & 1.00 & 0.92 & 1.00 & 0.89 & 0.87 & 0.53 & 0.70 & 0.56 & 0.95 & 0.85 \\
GPT-4-Turbo & 0.78 & 0.96 & 1.00 & 0.95 & 0.96 & 0.69 & 0.72 & 0.55 & 0.57 & 0.50 & 0.88 & 0.78 \\
GPT-4o & 0.84 & 0.97 & 1.00 & 0.90 & 0.96 & 0.74 & 0.81 & 0.51 & 0.38 & 0.50 & 0.91 & 0.77 \\
Deepseek-R1-70B & 0.74 & 0.87 & 1.00 & 0.82 & 0.89 & 0.52 & 0.66 & 0.45 & 0.33 & 0.31 & 0.81 & 0.67 \\
Llama-3-70B-Instruct & 0.83 & 0.89 & 0.95 & 0.79 & 0.94 & 0.39 & 0.66 & 0.38 & 0.46 & 0.25 & 0.82 & 0.67 \\
Gemma-2-27B-Instruct & 0.76 & 0.92 & 1.00 & 0.79 & 0.94 & 0.50 & 0.53 & 0.34 & 0.33 & 0.38 & 0.80 & 0.66 \\
Mixtral-8x22B-Instruct & 0.71 & 0.85 & 0.90 & 0.76 & 0.92 & 0.42 & 0.52 & 0.32 & 0.31 & 0.44 & 0.77 & 0.63 \\
Mixtral-8x7B-Instruct & 0.72 & 0.84 & 0.87 & 0.58 & 0.87 & 0.32 & 0.52 & 0.32 & 0.23 & 0.25 & 0.78 & 0.57 \\
GPT-3.5-Turbo & 0.64 & 0.80 & 0.90 & 0.53 & 0.87 & 0.32 & 0.44 & 0.20 & 0.20 & 0.38 & 0.74 & 0.55 \\
Llama-3-8B-Instruct & 0.57 & 0.71 & 0.82 & 0.56 & 0.79 & 0.26 & 0.37 & 0.22 & 0.30 & 0.25 & 0.67 & 0.50 \\
Mistral-7B-Instruct & 0.52 & 0.67 & 0.72 & 0.42 & 0.77 & 0.25 & 0.40 & 0.05 & 0.27 & 0.06 & 0.62 & 0.43 \\
Gemma-7B-Instruct & 0.40 & 0.63 & 0.77 & 0.32 & 0.66 & 0.12 & 0.36 & 0.12 & 0.14 & 0.06 & 0.58 & 0.38 \\
Llama-2-7B-Chat & 0.29 & 0.41 & 0.31 & 0.19 & 0.56 & 0.12 & 0.34 & 0.14 & 0.12 & -0.12 & 0.44 & 0.25 \\
Salamandra-7B-Instruct & 0.29 & 0.41 & 0.36 & 0.27 & 0.72 & 0.02 & 0.16 & -0.07 & -0.05 & -0.00 & 0.32 & 0.22 \\
Leniachat-Gemma-2B & 0.19 & 0.21 & 0.03 & 0.06 & 0.15 & 0.05 & 0.22 & -0.03 & 0.20 & -0.12 & 0.24 & 0.11 \\
\midrule
SPANISH-noto &&&&&&&&&&&&\\
\midrule
o3-mini & 0.47 & 0.91 & 0.89 & 0.82 & 0.79 & 0.43 & 0.08 & 0.98 & 0.86 & 1.00 & 0.58 & 0.71 \\
GPT-4o & 0.51 & 0.93 & 0.92 & 0.65 & 0.83 & 0.40 & 0.18 & 0.34 & 0.35 & 0.38 & 0.72 & 0.56 \\
Deepseek-R1-70B & 0.37 & 0.82 & 0.92 & 0.37 & 0.62 & 0.50 & 0.11 & 0.36 & 0.30 & 0.62 & 0.52 & 0.50 \\
GPT-4-Turbo & 0.41 & 0.83 & 0.92 & 0.58 & 0.77 & 0.39 & 0.09 & 0.38 & 0.23 & 0.19 & 0.67 & 0.50 \\
Claude-3.5-Sonnet & 0.39 & 0.75 & 0.92 & 0.58 & 0.81 & 0.40 & 0.24 & 0.18 & 0.14 & -0.00 & 0.59 & 0.45 \\
Llama-3-70B-Instruct & 0.03 & 0.41 & 0.65 & 0.30 & 0.37 & -0.05 & -0.20 & -0.09 & 0.01 & -0.06 & 0.44 & 0.16 \\
Gemma-2-27B-Instruct & 0.11 & 0.41 & 0.56 & 0.37 & 0.60 & 0.04 & -0.05 & -0.09 & -0.29 & -0.31 & 0.40 & 0.16 \\
Mistral-7B-Instruct & -0.03 & -0.00 & 0.26 & 0.19 & 0.30 & 0.11 & -0.19 & -0.07 & -0.09 & -0.06 & 0.23 & 0.06 \\
Llama-3-8B-Instruct & -0.20 & 0.05 & 0.15 & 0.09 & 0.22 & 0.22 & -0.25 & 0.10 & -0.10 & -0.00 & 0.22 & 0.05 \\
Mixtral-8x7B-Instruct & -0.18 & 0.05 & 0.15 & 0.09 & 0.32 & 0.16 & -0.25 & 0.16 & -0.12 & -0.12 & 0.20 & 0.04 \\
Mixtral-8x22B-Instruct & -0.22 & -0.00 & 0.18 & 0.12 & 0.22 & 0.02 & -0.22 & -0.17 & -0.28 & -0.38 & -0.05 & -0.07 \\
Salamandra-7B-Instruct & -0.29 & -0.25 & -0.20 & -0.05 & 0.01 & -0.22 & -0.22 & -0.21 & -0.32 & -0.38 & -0.07 & -0.20 \\
Llama-2-7B-Chat & -0.27 & -0.23 & -0.34 & -0.23 & -0.02 & 0.02 & -0.16 & -0.36 & -0.24 & -0.44 & -0.05 & -0.21 \\
GPT-3.5-Turbo & -0.48 & -0.37 & -0.20 & -0.30 & -0.08 & -0.01 & -0.29 & -0.42 & -0.40 & -0.44 & -0.20 & -0.29 \\
Gemma-7B-Instruct & -0.33 & -0.36 & -0.23 & -0.23 & -0.10 & -0.18 & -0.33 & -0.48 & -0.45 & -0.50 & -0.16 & -0.30 \\
Leniachat-Gemma-2B & -0.40 & -0.47 & -0.47 & -0.30 & -0.29 & -0.15 & -0.27 & -0.42 & -0.47 & -0.38 & -0.30 & -0.36 \\

\bottomrule
\end{tabular}
}

\caption{Cohen's Kappa results on \textbf{\textit{UNED-Access 2024}} by model and subject in English and Spanish, sorted by average.}
\label{tablauned}
\end{table*}

\begin{table*}[ht]
\centering
\resizebox{\linewidth}{!}{%
\renewcommand{\arraystretch}{0.8}
\begin{tabular}{lcccccccccccccccccc}

 & \textbf{\begin{sideways}abstract algebra\end{sideways}} & \textbf{\begin{sideways}anatomy\end{sideways}} & \textbf{\begin{sideways}astronomy\end{sideways}} & \textbf{\begin{sideways}college biology\end{sideways}} & \textbf{\begin{sideways}college chemistry\end{sideways}} & \textbf{\begin{sideways}college computer science\end{sideways}} & \textbf{\begin{sideways}college mathematics\end{sideways}} & \textbf{\begin{sideways}college physics\end{sideways}} & \textbf{\begin{sideways}computer security\end{sideways}} & \textbf{\begin{sideways}conceptual physics\end{sideways}} & \textbf{\begin{sideways}electrical engineering\end{sideways}} & \textbf{\begin{sideways}elementary mathematics\end{sideways}} &
 \textbf{\begin{sideways}high school biology\end{sideways}} &
 \textbf{\begin{sideways}high school chemistry\end{sideways}} &
 \textbf{\begin{sideways}high school computer science\end{sideways}} &
 \textbf{\begin{sideways}high school mathematics\end{sideways}} &
 \textbf{\begin{sideways}high school physics\end{sideways}} &
 \textbf{\begin{sideways}machine learning\end{sideways}}\\
\toprule
ENGLISH-original  &&&&&&&&&&&&\\
\midrule
o3-mini & 0.95 & 0.83 & 0.93 & 0.96 & 0.65 & 0.93 & 0.96 & 0.99 & 0.83 & 0.94 & 0.81 & 0.97 & 0.94 & 0.90 & 0.97 & 0.99 & 0.90 & 0.82 \\
Claude-3.5-Sonnet & 0.66 & 0.77 & 0.95 & 0.93 & 0.52 & 0.74 & 0.46 & 0.62 & 0.82 & 0.88 & 0.73 & 0.86 & 0.93 & 0.76 & 0.91 & 0.55 & 0.67 & 0.76 \\
GPT-4o & 0.42 & 0.85 & 0.94 & 0.93 & 0.43 & 0.74 & 0.25 & 0.54 & 0.82 & 0.86 & 0.75 & 0.68 & 0.93 & 0.72 & 0.90 & 0.34 & 0.67 & 0.67 \\
GPT-4-Turbo & 0.47 & 0.72 & 0.90 & 0.88 & 0.37 & 0.64 & 0.29 & 0.40 & 0.78 & 0.86 & 0.68 & 0.59 & 0.91 & 0.63 & 0.87 & 0.38 & 0.51 & 0.72 \\
Llama-3-70B-Instruct & 0.35 & 0.69 & 0.84 & 0.88 & 0.44 & 0.51 & 0.30 & 0.42 & 0.75 & 0.73 & 0.68 & 0.65 & 0.86 & 0.56 & 0.77 & 0.35 & 0.58 & 0.53 \\
Gemma-2-27B-Instruct & 0.26 & 0.69 & 0.80 & 0.88 & 0.31 & 0.57 & 0.29 & 0.36 & 0.71 & 0.68 & 0.60 & 0.44 & 0.91 & 0.59 & 0.76 & 0.28 & 0.45 & 0.44 \\
Mixtral-8x22B-Instruct & 0.27 & 0.64 & 0.77 & 0.82 & 0.35 & 0.51 & 0.14 & 0.37 & 0.66 & 0.63 & 0.54 & 0.45 & 0.82 & 0.55 & 0.77 & 0.21 & 0.38 & 0.33 \\
Deepseek-R1-70B & 0.26 & 0.49 & 0.77 & 0.79 & 0.36 & 0.36 & 0.26 & 0.29 & 0.64 & 0.75 & 0.56 & 0.41 & 0.81 & 0.42 & 0.63 & 0.20 & 0.30 & 0.25 \\
Mixtral-8x7B-Instruct & 0.19 & 0.55 & 0.73 & 0.79 & 0.33 & 0.49 & 0.14 & 0.29 & 0.72 & 0.50 & 0.47 & 0.26 & 0.75 & 0.42 & 0.57 & 0.17 & 0.29 & 0.33 \\
GPT-3.5-Turbo & 0.07 & 0.55 & 0.68 & 0.63 & 0.28 & 0.36 & 0.07 & 0.24 & 0.69 & 0.42 & 0.42 & 0.35 & 0.72 & 0.38 & 0.58 & 0.16 & 0.19 & 0.28 \\
Llama-3-8B-Instruct & 0.00 & 0.58 & 0.62 & 0.67 & 0.28 & 0.31 & 0.00 & 0.25 & 0.62 & 0.48 & 0.40 & 0.24 & 0.71 & 0.35 & 0.51 & 0.05 & 0.23 & 0.25 \\
Mistral-7B-Instruct & 0.10 & 0.41 & 0.52 & 0.51 & 0.17 & 0.43 & 0.12 & 0.18 & 0.59 & 0.32 & 0.37 & 0.16 & 0.58 & 0.26 & 0.43 & 0.12 & 0.11 & 0.29 \\
Gemma-7B-Instruct & -0.13 & 0.32 & 0.36 & 0.44 & 0.13 & 0.25 & -0.05 & -0.02 & 0.58 & 0.28 & 0.27 & -0.10 & 0.49 & 0.14 & 0.28 & -0.24 & -0.04 & 0.26 \\
Llama-2-7B-Chat & 0.15 & 0.27 & 0.20 & 0.21 & 0.03 & 0.10 & -0.02 & -0.14 & 0.31 & 0.14 & 0.10 & 0.08 & 0.23 & 0.08 & 0.13 & -0.05 & 0.00 & 0.17 \\
Salamandra-7B-Instruct & -0.09 & 0.14 & 0.12 & 0.23 & -0.05 & 0.06 & -0.09 & -0.06 & 0.28 & 0.13 & 0.16 & -0.08 & 0.24 & -0.04 & 0.11 & -0.09 & -0.06 & 0.07 \\
Leniachat-Gemma-2B & -0.05 & 0.17 & 0.09 & 0.07 & -0.03 & -0.04 & -0.05 & -0.01 & 0.27 & 0.08 & 0.07 & 0.02 & 0.13 & 0.06 & 0.08 & -0.03 & -0.04 & 0.13 \\
\midrule
ENGLISH-noto &&&&&&&&&&&&\\
\midrule
o3-mini & 0.88 & 0.55 & 0.64 & 0.74 & 0.48 & 0.74 & 0.91 & 0.96 & 0.41 & 0.62 & 0.48 & 0.93 & 0.60 & 0.64 & 0.74 & 0.96 & 0.76 & 0.68 \\
Deepseek-R1-70B & 0.26 & 0.39 & 0.41 & 0.57 & 0.53 & 0.32 & 0.56 & 0.70 & 0.23 & 0.51 & 0.28 & 0.64 & 0.45 & 0.63 & 0.41 & 0.70 & 0.72 & -0.03 \\
GPT-4o & 0.04 & 0.55 & 0.52 & 0.71 & 0.12 & 0.29 & 0.14 & 0.16 & 0.42 & 0.52 & 0.36 & 0.22 & 0.64 & 0.37 & 0.54 & -0.03 & 0.30 & 0.18 \\
GPT-4-Turbo & -0.16 & 0.27 & 0.44 & 0.49 & 0.12 & 0.24 & 0.11 & 0.16 & 0.30 & 0.29 & 0.10 & 0.15 & 0.52 & 0.23 & 0.40 & 0.16 & 0.11 & 0.08 \\
Claude-3.5-Sonnet & -0.12 & 0.26 & 0.27 & 0.36 & 0.03 & 0.24 & -0.04 & 0.15 & 0.05 & 0.19 & 0.04 & 0.07 & 0.41 & 0.08 & 0.28 & -0.15 & 0.03 & -0.01 \\
Gemma-2-27B-Instruct & -0.31 & 0.02 & 0.05 & 0.36 & -0.09 & -0.11 & -0.02 & -0.16 & 0.07 & 0.01 & -0.06 & -0.19 & 0.26 & 0.08 & 0.11 & -0.23 & -0.10 & -0.14 \\
Llama-3-70B-Instruct & -0.24 & 0.10 & 0.01 & 0.30 & -0.15 & 0.01 & 0.02 & -0.08 & 0.02 & -0.09 & -0.17 & -0.02 & 0.17 & -0.05 & 0.14 & -0.15 & -0.04 & -0.14 \\
Mixtral-8x7B-Instruct & -0.10 & -0.01 & 0.00 & 0.11 & -0.09 & 0.15 & -0.01 & -0.11 & 0.09 & -0.06 & 0.00 & -0.11 & 0.10 & -0.04 & 0.07 & -0.11 & 0.00 & -0.13 \\
Llama-2-7B-Chat & -0.12 & 0.05 & -0.05 & -0.02 & -0.11 & 0.08 & 0.06 & -0.15 & 0.09 & 0.12 & 0.01 & -0.03 & 0.00 & -0.02 & -0.09 & -0.08 & -0.13 & -0.07 \\
Llama-3-8B-Instruct & -0.21 & 0.06 & -0.06 & 0.12 & -0.12 & 0.01 & 0.07 & -0.05 & 0.06 & -0.10 & -0.14 & -0.17 & 0.05 & -0.09 & -0.02 & -0.04 & -0.02 & -0.21 \\
Mistral-7B-Instruct & -0.21 & -0.08 & -0.10 & 0.05 & -0.16 & -0.19 & -0.10 & -0.18 & -0.01 & -0.08 & -0.07 & -0.24 & -0.04 & -0.08 & -0.09 & -0.18 & -0.21 & -0.22 \\
Mixtral-8x22B-Instruct & -0.25 & -0.23 & -0.12 & 0.01 & -0.17 & -0.15 & -0.04 & -0.16 & -0.22 & -0.18 & -0.25 & -0.22 & -0.03 & -0.17 & -0.12 & -0.18 & -0.12 & -0.25 \\
Salamandra-7B-Instruct & -0.32 & -0.19 & -0.22 & -0.12 & -0.25 & -0.14 & -0.24 & -0.10 & -0.24 & -0.11 & -0.19 & -0.23 & -0.14 & -0.20 & -0.13 & -0.25 & -0.13 & -0.25 \\
GPT-3.5-Turbo & -0.28 & -0.24 & -0.24 & -0.13 & -0.25 & -0.26 & -0.27 & -0.29 & -0.21 & -0.30 & -0.30 & -0.29 & -0.16 & -0.22 & -0.23 & -0.27 & -0.29 & -0.25 \\
Gemma-7B-Instruct & -0.33 & -0.25 & -0.26 & -0.19 & -0.23 & -0.24 & -0.28 & -0.28 & -0.25 & -0.28 & -0.27 & -0.30 & -0.20 & -0.27 & -0.28 & -0.29 & -0.30 & -0.31 \\
Leniachat-Gemma-2B & -0.33 & -0.30 & -0.32 & -0.31 & -0.32 & -0.28 & -0.29 & -0.29 & -0.32 & -0.32 & -0.32 & -0.30 & -0.28 & -0.29 & -0.29 & -0.26 & -0.30 & -0.31 \\
\midrule
SPANISH-original  &&&&&&&&&&&&\\
\midrule
o3-mini & 0.93 & 0.82 & 0.94 & 0.94 & 0.64 & 0.96 & 0.96 & 0.97 & 0.80 & 0.92 & 0.78 & 0.95 & 0.93 & 0.88 & 0.96 & 0.98 & 0.86 & 0.81 \\
Claude-3.5-Sonnet & 0.66 & 0.77 & 0.93 & 0.92 & 0.47 & 0.75 & 0.42 & 0.66 & 0.78 & 0.89 & 0.69 & 0.89 & 0.91 & 0.76 & 0.93 & 0.49 & 0.66 & 0.69 \\
GPT-4o & 0.45 & 0.79 & 0.93 & 0.90 & 0.40 & 0.69 & 0.34 & 0.48 & 0.79 & 0.87 & 0.72 & 0.69 & 0.93 & 0.69 & 0.83 & 0.33 & 0.56 & 0.67 \\
GPT-4-Turbo & 0.46 & 0.63 & 0.89 & 0.87 & 0.31 & 0.65 & 0.26 & 0.37 & 0.80 & 0.81 & 0.64 & 0.57 & 0.90 & 0.61 & 0.84 & 0.32 & 0.49 & 0.60 \\
Llama-3-70B-Instruct & 0.29 & 0.54 & 0.82 & 0.83 & 0.41 & 0.54 & 0.19 & 0.41 & 0.73 & 0.68 & 0.67 & 0.60 & 0.85 & 0.51 & 0.77 & 0.33 & 0.46 & 0.46 \\
Gemma-2-27B-Instruct & 0.23 & 0.58 & 0.75 & 0.86 & 0.33 & 0.58 & 0.27 & 0.33 & 0.68 & 0.62 & 0.57 & 0.41 & 0.88 & 0.56 & 0.73 & 0.24 & 0.45 & 0.47 \\
Deepseek-R1-70B & 0.19 & 0.56 & 0.77 & 0.75 & 0.33 & 0.42 & 0.22 & 0.32 & 0.75 & 0.64 & 0.53 & 0.41 & 0.78 & 0.51 & 0.73 & 0.27 & 0.32 & 0.49 \\
Mixtral-8x22B-Instruct & 0.25 & 0.46 & 0.68 & 0.70 & 0.28 & 0.56 & 0.16 & 0.29 & 0.68 & 0.60 & 0.44 & 0.42 & 0.77 & 0.45 & 0.74 & 0.25 & 0.31 & 0.37 \\
Mixtral-8x7B-Instruct & 0.10 & 0.43 & 0.68 & 0.62 & 0.33 & 0.37 & 0.06 & 0.16 & 0.72 & 0.46 & 0.37 & 0.30 & 0.72 & 0.33 & 0.54 & 0.20 & 0.28 & 0.19 \\
GPT-3.5-Turbo & 0.08 & 0.45 & 0.59 & 0.52 & 0.24 & 0.35 & 0.22 & 0.19 & 0.65 & 0.40 & 0.41 & 0.25 & 0.65 & 0.32 & 0.57 & 0.07 & 0.08 & 0.13 \\
Llama-3-8B-Instruct & 0.00 & 0.35 & 0.53 & 0.42 & 0.17 & 0.25 & -0.04 & 0.03 & 0.55 & 0.30 & 0.37 & 0.16 & 0.67 & 0.24 & 0.48 & 0.01 & 0.16 & 0.18 \\
Mistral-7B-Instruct & 0.03 & 0.21 & 0.34 & 0.40 & 0.16 & 0.24 & 0.04 & 0.12 & 0.47 & 0.23 & 0.30 & 0.15 & 0.49 & 0.28 & 0.37 & 0.09 & 0.11 & 0.31 \\
Gemma-7B-Instruct & -0.10 & 0.23 & 0.30 & 0.28 & 0.19 & 0.19 & 0.02 & -0.02 & 0.52 & 0.24 & 0.25 & -0.01 & 0.39 & 0.18 & 0.27 & -0.10 & 0.00 & 0.15 \\
Llama-2-7B-Chat & -0.06 & 0.13 & 0.16 & 0.14 & -0.08 & 0.08 & 0.00 & -0.18 & 0.21 & 0.15 & 0.16 & 0.09 & 0.23 & 0.03 & 0.13 & 0.04 & 0.01 & 0.15 \\
Salamandra-7B-Instruct & -0.09 & -0.03 & 0.11 & 0.18 & -0.03 & 0.11 & -0.04 & -0.20 & 0.24 & 0.11 & 0.14 & -0.01 & 0.19 & -0.08 & 0.08 & -0.10 & -0.02 & 0.10 \\
Leniachat-Gemma-2B & 0.02 & 0.09 & 0.10 & -0.00 & 0.07 & 0.01 & 0.04 & -0.05 & 0.16 & 0.08 & 0.10 & -0.01 & 0.16 & 0.11 & 0.11 & -0.05 & -0.05 & 0.04 \\
\midrule
SPANISH-noto &&&&&&&&&&&&\\
\midrule
o3-mini & 0.87 & 0.51 & 0.57 & 0.60 & 0.35 & 0.63 & 0.91 & 0.93 & 0.35 & 0.66 & 0.35 & 0.91 & 0.58 & 0.61 & 0.73 & 0.94 & 0.72 & 0.61 \\
GPT-4o & 0.04 & 0.48 & 0.54 & 0.61 & 0.20 & 0.28 & 0.15 & 0.15 & 0.37 & 0.55 & 0.31 & 0.25 & 0.63 & 0.33 & 0.56 & -0.01 & 0.21 & 0.25 \\
GPT-4-Turbo & -0.13 & 0.26 & 0.47 & 0.49 & 0.23 & 0.22 & 0.30 & 0.23 & 0.27 & 0.35 & 0.10 & 0.18 & 0.55 & 0.28 & 0.40 & 0.25 & 0.17 & 0.08 \\
Deepseek-R1-70B & -0.01 & 0.23 & 0.36 & 0.20 & 0.23 & 0.19 & 0.29 & 0.56 & 0.14 & 0.29 & 0.10 & 0.31 & 0.27 & 0.45 & 0.28 & 0.38 & 0.53 & -0.01 \\
Claude-3.5-Sonnet & -0.10 & 0.30 & 0.27 & 0.45 & 0.00 & 0.29 & -0.02 & 0.06 & 0.13 & 0.30 & -0.02 & 0.10 & 0.43 & 0.15 & 0.38 & -0.13 & 0.13 & -0.03 \\
Gemma-2-27B-Instruct & -0.32 & 0.00 & 0.05 & 0.29 & -0.16 & -0.07 & -0.08 & -0.12 & 0.00 & 0.01 & -0.09 & -0.23 & 0.19 & 0.03 & 0.10 & -0.23 & -0.10 & -0.15 \\
Llama-3-70B-Instruct & -0.21 & -0.04 & -0.07 & 0.08 & -0.20 & -0.11 & -0.08 & -0.18 & 0.05 & -0.03 & -0.20 & -0.06 & 0.10 & -0.06 & 0.10 & -0.18 & -0.09 & -0.18 \\
Llama-3-8B-Instruct & -0.21 & -0.11 & -0.11 & 0.01 & -0.11 & 0.01 & 0.02 & -0.03 & -0.04 & -0.05 & -0.17 & -0.19 & -0.02 & -0.01 & 0.07 & -0.13 & 0.01 & -0.11 \\
Mixtral-8x7B-Instruct & -0.13 & -0.08 & -0.07 & 0.00 & -0.04 & -0.03 & 0.00 & -0.16 & -0.08 & -0.13 & -0.10 & -0.16 & 0.00 & -0.04 & -0.00 & -0.11 & -0.09 & -0.17 \\
Llama-2-7B-Chat & -0.19 & -0.02 & -0.11 & -0.11 & -0.23 & 0.04 & 0.03 & -0.14 & -0.07 & -0.15 & -0.07 & -0.08 & -0.14 & -0.09 & -0.13 & -0.09 & -0.15 & -0.11 \\
Mistral-7B-Instruct & -0.17 & -0.15 & -0.10 & -0.01 & -0.17 & -0.10 & 0.04 & -0.15 & -0.02 & -0.12 & -0.04 & -0.24 & -0.04 & -0.08 & -0.08 & -0.10 & -0.20 & -0.06 \\
Mixtral-8x22B-Instruct & -0.28 & -0.17 & -0.13 & -0.08 & -0.13 & -0.14 & -0.04 & -0.27 & -0.21 & -0.17 & -0.20 & -0.23 & -0.05 & -0.14 & -0.05 & -0.23 & -0.15 & -0.26 \\
Salamandra-7B-Instruct & -0.24 & -0.26 & -0.17 & -0.21 & -0.16 & -0.03 & -0.02 & -0.20 & -0.09 & -0.08 & -0.17 & -0.14 & -0.16 & -0.19 & -0.06 & -0.13 & -0.17 & -0.10 \\
GPT-3.5-Turbo & -0.27 & -0.30 & -0.20 & -0.14 & -0.25 & -0.24 & -0.24 & -0.25 & -0.28 & -0.29 & -0.32 & -0.28 & -0.18 & -0.25 & -0.18 & -0.20 & -0.28 & -0.28 \\
Gemma-7B-Instruct & -0.32 & -0.20 & -0.20 & -0.21 & -0.27 & -0.22 & -0.29 & -0.31 & -0.26 & -0.27 & -0.25 & -0.30 & -0.24 & -0.23 & -0.23 & -0.32 & -0.30 & -0.29 \\
Leniachat-Gemma-2B & -0.32 & -0.33 & -0.30 & -0.32 & -0.32 & -0.32 & -0.28 & -0.31 & -0.33 & -0.33 & -0.28 & -0.29 & -0.30 & -0.31 & -0.32 & -0.26 & -0.33 & -0.33 \\

\bottomrule
\end{tabular}
}

\caption{Cohen's Kappa results on \textbf{MMLU} (\textbf{\textcolor{blue}{STEM}}) by model and subject in English and Spanish, sorted by average.}
\label{tablammlu1}
\end{table*}

\begin{table*}[ht]
\centering
\scriptsize
\resizebox{\linewidth}{!}{%
\renewcommand{\arraystretch}{0.8}
\begin{tabular}{lccccccccccc}

 & \textbf{\begin{sideways}econometrics\end{sideways}} & \textbf{\begin{sideways}high school geography\end{sideways}} & \textbf{\tiny\begin{sideways}h.s. government and politics\end{sideways}} & \textbf{\tiny\begin{sideways}h.s. macroeconomics\end{sideways}} & \textbf{\tiny\begin{sideways}h.s. microeconomics\end{sideways}} & \textbf{\begin{sideways}high school psychology\end{sideways}} & \textbf{\begin{sideways}high school statistics\end{sideways}} & \textbf{\begin{sideways}public relations\end{sideways}} & \textbf{\begin{sideways}security studies\end{sideways}} & \textbf{\begin{sideways}sociology\end{sideways}} & \textbf{\begin{sideways}us foreign policy\end{sideways}} \\
\toprule
ENGLISH-original  &&&&&&&&&&&\\
\midrule
o3-mini & 0.83 & 0.91 & 0.97 & 0.91 & 0.96 & 0.95 & 0.91 & 0.67 & 0.72 & 0.89 & 0.89 \\
Claude-3.5-Sonnet & 0.72 & 0.90 & 0.96 & 0.86 & 0.97 & 0.93 & 0.78 & 0.77 & 0.78 & 0.93 & 0.93 \\
GPT-4o & 0.58 & 0.91 & 0.97 & 0.88 & 0.97 & 0.94 & 0.70 & 0.69 & 0.78 & 0.94 & 0.93 \\
GPT-4-Turbo & 0.58 & 0.93 & 0.97 & 0.82 & 0.93 & 0.93 & 0.70 & 0.70 & 0.69 & 0.88 & 0.93 \\
Llama-3-70B-Instruct & 0.58 & 0.88 & 0.97 & 0.77 & 0.83 & 0.92 & 0.63 & 0.64 & 0.71 & 0.86 & 0.89 \\
Gemma-2-27B-Instruct & 0.50 & 0.90 & 0.99 & 0.79 & 0.84 & 0.91 & 0.59 & 0.64 & 0.70 & 0.85 & 0.87 \\
Mixtral-8x22B-Instruct & 0.49 & 0.79 & 0.94 & 0.67 & 0.78 & 0.87 & 0.60 & 0.70 & 0.70 & 0.88 & 0.91\\
Deepseek-R1-70B & 0.46 & 0.78 & 0.91 & 0.72 & 0.79 & 0.89 & 0.59 & 0.62 & 0.66 & 0.85 & 0.78 \\
Mixtral-8x7B-Instruct & 0.50 & 0.79 & 0.92 & 0.55 & 0.63 & 0.81 & 0.42 & 0.63 & 0.66 & 0.83 & 0.84 \\
GPT-3.5-Turbo & 0.29 & 0.76 & 0.87 & 0.56 & 0.63 & 0.81 & 0.29 & 0.62 & 0.58 & 0.79 & 0.78 \\
Llama-3-8B-Instruct & 0.43 & 0.69 & 0.82 & 0.53 & 0.65 & 0.79 & 0.43 & 0.57 & 0.61 & 0.82 & 0.84 \\
Mistral-7B-Instruct & 0.28 & 0.61 & 0.77 & 0.37 & 0.49 & 0.71 & 0.27 & 0.57 & 0.49 & 0.81 & 0.71 \\
Gemma-7B-Instruct & 0.09 & 0.59 & 0.59 & 0.33 & 0.34 & 0.63 & -0.01 & 0.59 & 0.43 & 0.61 & 0.60 \\
Llama-2-7B-Chat & 0.06 & 0.32 & 0.33 & 0.11 & 0.10 & 0.37 & -0.04 & 0.36 & 0.15 & 0.33 & 0.57 \\
Salamandra-7B-Instruct & -0.00 & 0.25 & 0.38 & 0.15 & 0.11 & 0.42 & -0.04 & 0.22 & 0.17 & 0.51 & 0.44 \\
Leniachat-Gemma-2B & 0.01 & 0.17 & 0.21 & 0.08 & 0.08 & 0.21 & -0.02 & 0.19 & 0.10 & 0.20 & 0.21 \\
\midrule
ENGLISH-noto &&&&&&&&&&&\\
\midrule
o3-mini & 0.50 & 0.49 & 0.72 & 0.76 & 0.81 & 0.63 & 0.72 & 0.11 & 0.12 & 0.43 & 0.55 \\
Deepseek-R1-70B & 0.40 & 0.59 & 0.75 & 0.57 & 0.59 & 0.59 & 0.59 & 0.33 & 0.50 & 0.51 & 0.42 \\
GPT-4o & 0.36 & 0.63 & 0.80 & 0.60 & 0.77 & 0.67 & 0.30 & 0.16 & 0.26 & 0.52 & 0.73 \\
GPT-4-Turbo & 0.20 & 0.59 & 0.82 & 0.38 & 0.54 & 0.55 & 0.30 & 0.15 & 0.33 & 0.53 & 0.62 \\
Claude-3.5-Sonnet & 0.08 & 0.38 & 0.71 & 0.36 & 0.51 & 0.28 & 0.19 & -0.14 & 0.15 & 0.26 & 0.42 \\
Gemma-2-27B-Instruct & -0.11 & 0.28 & 0.56 & 0.15 & 0.27 & 0.38 & -0.03 & -0.06 & 0.21 & 0.30 & 0.30 \\
Llama-3-70B-Instruct & -0.08 & 0.23 & 0.46 & 0.12 & 0.24 & 0.23 & -0.01 & -0.09 & 0.06 & 0.03 & 0.24 \\
Mixtral-8x7B-Instruct & -0.11 & 0.38 & 0.40 & -0.02 & 0.08 & 0.21 & -0.10 & 0.10 & 0.04 & 0.13 & 0.23 \\
Llama-2-7B-Chat & -0.06 & 0.25 & 0.14 & -0.08 & 0.00 & 0.19 & -0.12 & 0.23 & -0.07 & 0.17 & 0.10 \\
Llama-3-8B-Instruct & -0.13 & 0.21 & 0.35 & -0.00 & 0.05 & 0.12 & -0.06 & 0.04 & 0.00 & 0.08 & 0.05 \\
Mistral-7B-Instruct & -0.17 & 0.06 & 0.07 & -0.11 & -0.05 & 0.02 & -0.20 & -0.07 & -0.10 & -0.01 & -0.08 \\
Mixtral-8x22B-Instruct & -0.21 & 0.04 & 0.14 & -0.10 & -0.07 & -0.03 & -0.08 & -0.21 & -0.01 & -0.09 & -0.14 \\
Salamandra-7B-Instruct & -0.20 & -0.23 & -0.24 & -0.23 & -0.15 & -0.19 & -0.22 & -0.25 & -0.19 & -0.14 & -0.17 \\
GPT-3.5-Turbo & -0.27 & -0.04 & -0.03 & -0.24 & -0.15 & -0.18 & -0.26 & -0.25 & -0.05 & -0.20 & -0.24 \\
Gemma-7B-Instruct & -0.25 & -0.11 & -0.20 & -0.27 & -0.25 & -0.22 & -0.28 & -0.21 & -0.21 & -0.26 & -0.26 \\
Leniachat-Gemma-2B & -0.32 & -0.30 & -0.33 & -0.32 & -0.32 & -0.30 & -0.29 & -0.30 & -0.30 & -0.30 & -0.30 \\
\midrule
SPANISH-original  &&&&&&&&&&&\\
\midrule
o3-mini & 0.76 & 0.86 & 0.89 & 0.89 & 0.95 & 0.93 & 0.90 & 0.58 & 0.72 & 0.85 & 0.91 \\
Claude-3.5-Sonnet & 0.76 & 0.94 & 0.93 & 0.83 & 0.92 & 0.93 & 0.78 & 0.72 & 0.78 & 0.92 & 0.95 \\
GPT-4o & 0.61 & 0.90 & 0.95 & 0.87 & 0.94 & 0.94 & 0.72 & 0.73 & 0.73 & 0.85 & 0.91 \\
GPT-4-Turbo & 0.58 & 0.86 & 0.92 & 0.75 & 0.92 & 0.89 & 0.65 & 0.63 & 0.66 & 0.83 & 0.91 \\
Llama-3-70B-Instruct & 0.55 & 0.85 & 0.87 & 0.67 & 0.75 & 0.86 & 0.60 & 0.63 & 0.64 & 0.82 & 0.86 \\
Gemma-2-27B-Instruct & 0.46 & 0.84 & 0.88 & 0.72 & 0.82 & 0.87 & 0.61 & 0.63 & 0.71 & 0.81 & 0.84 \\
Deepseek-R1-70B & 0.47 & 0.82 & 0.82 & 0.67 & 0.72 & 0.83 & 0.56 & 0.62 & 0.70 & 0.80 & 0.84 \\
Mixtral-8x22B-Instruct & 0.41 & 0.74 & 0.77 & 0.61 & 0.74 & 0.84 & 0.56 & 0.57 & 0.70 & 0.77 & 0.75 \\
Mixtral-8x7B-Instruct & 0.39 & 0.75 & 0.73 & 0.51 & 0.61 & 0.72 & 0.34 & 0.62 & 0.68 & 0.74 & 0.82 \\
GPT-3.5-Turbo & 0.22 & 0.65 & 0.67 & 0.44 & 0.50 & 0.70 & 0.23 & 0.57 & 0.49 & 0.62 & 0.66 \\
Llama-3-8B-Instruct & 0.17 & 0.61 & 0.64 & 0.44 & 0.47 & 0.63 & 0.30 & 0.42 & 0.54 & 0.70 & 0.62 \\
Mistral-7B-Instruct & 0.18 & 0.53 & 0.45 & 0.23 & 0.35 & 0.54 & 0.22 & 0.40 & 0.42 & 0.63 & 0.68 \\
Gemma-7B-Instruct & 0.09 & 0.44 & 0.38 & 0.25 & 0.26 & 0.50 & 0.01 & 0.35 & 0.33 & 0.47 & 0.57 \\
Llama-2-7B-Chat & 0.03 & 0.28 & 0.22 & 0.05 & 0.06 & 0.27 & -0.03 & 0.32 & 0.23 & 0.33 & 0.51 \\
Salamandra-7B-Instruct & -0.04 & 0.26 & 0.35 & 0.10 & 0.11 & 0.33 & -0.04 & 0.31 & 0.06 & 0.38 & 0.21 \\
Leniachat-Gemma-2B & -0.04 & 0.07 & 0.08 & 0.08 & 0.10 & 0.19 & 0.05 & 0.14 & 0.04 & 0.14 & 0.15 \\
\midrule
SPANISH-noto &&&&&&&&&&&\\
\midrule
o3-mini & 0.46 & 0.47 & 0.61 & 0.67 & 0.69 & 0.57 & 0.62 & 0.00 & 0.03 & 0.30 & 0.39 \\
GPT-4o & 0.31 & 0.60 & 0.73 & 0.51 & 0.68 & 0.62 & 0.30 & 0.20 & 0.25 & 0.54 & 0.55 \\
GPT-4-Turbo & 0.22 & 0.61 & 0.80 & 0.45 & 0.62 & 0.56 & 0.32 & 0.21 & 0.36 & 0.54 & 0.60 \\
Deepseek-R1-70B & 0.16 & 0.26 & 0.45 & 0.32 & 0.37 & 0.32 & 0.33 & -0.04 & 0.18 & 0.23 & 0.44 \\
Claude-3.5-Sonnet & 0.23 & 0.46 & 0.63 & 0.38 & 0.49 & 0.39 & 0.19 & -0.06 & 0.20 & 0.36 & 0.42 \\
Gemma-2-27B-Instruct & -0.16 & 0.21 & 0.34 & 0.08 & 0.16 & 0.21 & -0.08 & -0.09 & 0.16 & 0.27 & 0.19 \\
Llama-3-70B-Instruct & -0.15 & 0.13 & 0.28 & 0.03 & 0.12 & 0.17 & -0.07 & -0.09 & 0.01 & -0.01 & 0.10 \\
Llama-3-8B-Instruct & -0.11 & 0.21 & 0.09 & -0.05 & -0.02 & 0.06 & 0.01 & -0.01 & -0.04 & -0.03 & -0.03 \\
Mixtral-8x7B-Instruct & -0.21 & 0.23 & 0.13 & -0.12 & -0.06 & -0.07 & -0.18 & 0.02 & 0.01 & 0.04 & -0.06 \\
Llama-2-7B-Chat & -0.08 & 0.12 & -0.09 & -0.19 & -0.14 & -0.02 & -0.13 & 0.16 & -0.16 & -0.14 & 0.05 \\
Mistral-7B-Instruct & -0.10 & 0.15 & -0.00 & -0.16 & -0.05 & -0.01 & -0.12 & -0.07 & -0.11 & -0.07 & -0.12 \\
Mixtral-8x22B-Instruct & -0.24 & 0.05 & 0.14 & -0.11 & -0.08 & -0.06 & -0.08 & -0.16 & -0.02 & -0.09 & -0.12 \\
Salamandra-7B-Instruct & -0.10 & -0.04 & -0.03 & -0.22 & -0.20 & -0.21 & -0.20 & -0.20 & -0.21 & -0.13 & -0.23 \\
GPT-3.5-Turbo & -0.22 & -0.01 & -0.14 & -0.27 & -0.21 & -0.24 & -0.24 & -0.20 & -0.12 & -0.25 & -0.30 \\
Gemma-7B-Instruct & -0.30 & -0.10 & -0.19 & -0.28 & -0.26 & -0.20 & -0.29 & -0.16 & -0.26 & -0.26 & -0.15 \\
Leniachat-Gemma-2B & -0.32 & -0.27 & -0.26 & -0.31 & -0.30 & -0.27 & -0.29 & -0.33 & -0.33 & -0.30 & -0.32 \\
\bottomrule
\end{tabular}
}

\caption{Cohen's Kappa results on \textbf{MMLU} (\textbf{\textcolor{blue}{Social Sciences}}) by model and subject in English and Spanish, sorted by average.}
\label{tablammlu2}
\end{table*}

\begin{table*}[ht]
\centering
\resizebox{\linewidth}{!}{%
\renewcommand{\arraystretch}{0.8}
\begin{tabular}{lccccccccccccc}

 & \textbf{\begin{sideways}formal logic\end{sideways}} & \textbf{\small\begin{sideways}h.s. european history\end{sideways}} & \textbf{\begin{sideways}h.s. us history\end{sideways}} & \textbf{\begin{sideways}h.s. world history\end{sideways}} & \textbf{\begin{sideways}international law\end{sideways}} & 
 \textbf{\begin{sideways}jurisprudence\end{sideways}} &
 \textbf{\begin{sideways}logical fallacies\end{sideways}} & \textbf{\begin{sideways}moral disputes\end{sideways}} & \textbf{\begin{sideways}moral scenarios\end{sideways}} & \textbf{\begin{sideways}philosophy\end{sideways}} & \textbf{\begin{sideways}prehistory\end{sideways}} & \textbf{\begin{sideways}professional law\end{sideways}} &
 \textbf{\begin{sideways}world religions\end{sideways}}\\
\toprule
ENGLISH-original  &&&&&&&&&&&\\
\midrule
o3-mini & 0.95 & 0.80 & 0.85 & 0.87 & 0.87 & 0.85 & 0.87 & 0.76 & 0.53 & 0.78 & 0.86 & 0.50 & 0.83 \\
Claude-3.5-Sonnet & 0.65 & 0.87 & 0.90 & 0.93 & 0.91 & 0.83 & 0.89 & 0.83 & 0.81 & 0.87 & 0.93 & 0.65 & 0.87 \\
GPT-4o & 0.56 & 0.86 & 0.92 & 0.93 & 0.89 & 0.86 & 0.86 & 0.84 & 0.56 & 0.91 & 0.92 & 0.66 & 0.87 \\
GPT-4-Turbo & 0.51 & 0.84 & 0.93 & 0.90 & 0.88 & 0.81 & 0.83 & 0.77 & 0.59 & 0.84 & 0.85 & 0.55 & 0.81 \\
Llama-3-70B-Instruct & 0.48 & 0.76 & 0.87 & 0.82 & 0.87 & 0.83 & 0.77 & 0.76 & 0.47 & 0.74 & 0.84 & 0.49 & 0.86 \\
Gemma-2-27B-Instruct & 0.40 & 0.79 & 0.86 & 0.89 & 0.79 & 0.80 & 0.77 & 0.70 & 0.31 & 0.77 & 0.79 & 0.46 & 0.83 \\
Mixtral-8x22B-Instruct & 0.48 & 0.79 & 0.81 & 0.81 & 0.88 & 0.76 & 0.76 & 0.68 & 0.40 & 0.77 & 0.76 & 0.44 & 0.77 \\
Deepseek-R1-70B & 0.33 & 0.76 & 0.87 & 0.81 & 0.82 & 0.77 & 0.68 & 0.61 & 0.72 & 0.75 & 0.76 & 0.48 & 0.79 \\
Mixtral-8x7B-Instruct & 0.29 & 0.73 & 0.77 & 0.81 & 0.79 & 0.76 & 0.69 & 0.66 & 0.07 & 0.69 & 0.74 & 0.35 & 0.83 \\
GPT-3.5-Turbo & 0.24 & 0.65 & 0.76 & 0.75 & 0.76 & 0.62 & 0.72 & 0.61 & 0.10 & 0.71 & 0.65 & 0.32 & 0.79 \\
Llama-3-8B-Instruct & 0.31 & 0.62 & 0.70 & 0.75 & 0.69 & 0.66 & 0.65 & 0.58 & -0.02 & 0.65 & 0.63 & 0.29 & 0.71 \\
Mistral-7B-Instruct & 0.20 & 0.64 & 0.67 & 0.66 & 0.65 & 0.62 & 0.67 & 0.55 & 0.15 & 0.50 & 0.54 & 0.22 & 0.74 \\
Gemma-7B-Instruct & 0.04 & 0.44 & 0.41 & 0.50 & 0.54 & 0.50 & 0.43 & 0.39 & -0.09 & 0.47 & 0.43 & 0.10 & 0.54 \\
Llama-2-7B-Chat & -0.01 & 0.28 & 0.20 & 0.28 & 0.43 & 0.29 & 0.30 & 0.20 & -0.01 & 0.24 & 0.35 & 0.10 & 0.49 \\
Salamandra-7B-Instruct & 0.06 & 0.42 & 0.44 & 0.46 & 0.35 & 0.25 & 0.29 & 0.21 & -0.25 & 0.21 & 0.21 & 0.07 & 0.38 \\
Leniachat-Gemma-2B & 0.14 & 0.10 & 0.12 & 0.07 & 0.19 & 0.07 & 0.17 & 0.07 & -0.01 & 0.13 & 0.11 & 0.06 & 0.23 \\
\midrule
ENGLISH-noto &&&&&&&&&&&\\
\midrule
o3-mini & 0.70 & 0.41 & 0.64 & 0.51 & 0.37 & 0.37 & 0.63 & 0.10 & 0.39 & 0.33 & 0.39 & -0.07 & 0.56 \\
Deepseek-R1-70B & 0.27 & 0.56 & 0.65 & 0.52 & 0.50 & 0.52 & 0.39 & 0.24 & 0.42 & 0.49 & 0.42 & 0.35 & 0.46 \\
GPT-4o & 0.24 & 0.47 & 0.72 & 0.61 & 0.48 & 0.64 & 0.74 & 0.52 & 0.18 & 0.56 & 0.58 & 0.09 & 0.65 \\
GPT-4-Turbo & 0.06 & 0.44 & 0.68 & 0.58 & 0.42 & 0.59 & 0.69 & 0.41 & -0.18 & 0.39 & 0.47 & -0.04 & 0.59 \\
Claude-3.5-Sonnet & -0.02 & 0.33 & 0.62 & 0.49 & 0.33 & 0.34 & 0.35 & 0.21 & -0.02 & 0.24 & 0.31 & 0.07 & 0.34 \\
Gemma-2-27B-Instruct & -0.22 & 0.30 & 0.51 & 0.41 & 0.15 & 0.28 & 0.34 & 0.14 & -0.31 & 0.18 & 0.11 & -0.12 & 0.13 \\
Llama-3-70B-Instruct & -0.14 & 0.28 & 0.45 & 0.30 & 0.06 & 0.09 & 0.27 & -0.09 & -0.11 & 0.01 & -0.04 & -0.06 & 0.15 \\
Mixtral-8x7B-Instruct & -0.18 & 0.19 & 0.44 & 0.24 & -0.04 & 0.09 & 0.27 & 0.10 & -0.26 & 0.08 & 0.08 & -0.14 & 0.26 \\
Llama-2-7B-Chat & -0.23 & 0.33 & 0.28 & 0.31 & 0.14 & 0.17 & 0.23 & 0.07 & -0.01 & 0.10 & 0.06 & -0.23 & 0.13 \\
Llama-3-8B-Instruct & -0.19 & 0.20 & 0.39 & 0.25 & -0.06 & 0.14 & 0.39 & -0.02 & -0.33 & 0.12 & 0.02 & -0.18 & 0.13 \\
Mistral-7B-Instruct & -0.25 & -0.04 & 0.19 & 0.10 & -0.20 & -0.09 & 0.18 & -0.08 & -0.27 & -0.10 & -0.11 & -0.28 & 0.03 \\
Mixtral-8x22B-Instruct & -0.19 & -0.07 & 0.06 & -0.11 & -0.29 & -0.10 & 0.01 & -0.12 & -0.14 & -0.05 & -0.12 & -0.24 & -0.13 \\
Salamandra-7B-Instruct & -0.26 & -0.05 & -0.06 & 0.00 & -0.23 & -0.26 & -0.25 & -0.19 & -0.26 & -0.19 & -0.19 & -0.21 & -0.11 \\
GPT-3.5-Turbo & -0.30 & -0.16 & 0.06 & -0.08 & -0.27 & -0.16 & -0.20 & -0.21 & -0.33 & -0.21 & -0.23 & -0.30 & -0.24 \\
Gemma-7B-Instruct & -0.28 & -0.19 & -0.12 & -0.12 & -0.32 & -0.26 & -0.29 & -0.24 & -0.15 & -0.30 & -0.26 & -0.29 & -0.27 \\
Leniachat-Gemma-2B & -0.27 & -0.18 & -0.16 & -0.20 & -0.31 & -0.33 & -0.28 & -0.31 & -0.26 & -0.28 & -0.33 & -0.28 & -0.30 \\
\midrule
SPANISH-original  &&&&&&&&&&&\\
\midrule
o3-mini & 0.88 & 0.77 & 0.79 & 0.84 & 0.81 & 0.76 & 0.80 & 0.70 & 0.40 & 0.75 & 0.81 & 0.40 & 0.83 \\
Claude-3.5-Sonnet & 0.55 & 0.84 & 0.90 & 0.90 & 0.88 & 0.85 & 0.87 & 0.82 & 0.73 & 0.90 & 0.90 & 0.58 & 0.87 \\
GPT-4o & 0.52 & 0.83 & 0.91 & 0.93 & 0.87 & 0.82 & 0.84 & 0.82 & 0.58 & 0.87 & 0.90 & 0.55 & 0.86 \\
GPT-4-Turbo & 0.46 & 0.82 & 0.85 & 0.86 & 0.85 & 0.80 & 0.84 & 0.76 & 0.54 & 0.78 & 0.83 & 0.46 & 0.82 \\
Llama-3-70B-Instruct & 0.43 & 0.79 & 0.78 & 0.79 & 0.78 & 0.77 & 0.74 & 0.67 & 0.24 & 0.68 & 0.79 & 0.37 & 0.80 \\
Gemma-2-27B-Instruct & 0.38 & 0.76 & 0.84 & 0.81 & 0.75 & 0.75 & 0.71 & 0.65 & 0.01 & 0.71 & 0.73 & 0.37 & 0.77 \\
Deepseek-R1-70B & 0.41 & 0.80 & 0.85 & 0.85 & 0.80 & 0.71 & 0.74 & 0.69 & 0.52 & 0.71 & 0.78 & 0.39 & 0.82 \\
Mixtral-8x22B-Instruct & 0.34 & 0.70 & 0.72 & 0.76 & 0.73 & 0.67 & 0.67 & 0.64 & 0.02 & 0.64 & 0.65 & 0.34 & 0.72 \\
Mixtral-8x7B-Instruct & 0.26 & 0.74 & 0.72 & 0.76 & 0.72 & 0.67 & 0.65 & 0.56 & 0.05 & 0.64 & 0.61 & 0.27 & 0.77 \\
GPT-3.5-Turbo & 0.20 & 0.65 & 0.66 & 0.72 & 0.65 & 0.61 & 0.63 & 0.53 & 0.03 & 0.59 & 0.53 & 0.24 & 0.71 \\
Llama-3-8B-Instruct & 0.20 & 0.63 & 0.60 & 0.71 & 0.63 & 0.64 & 0.47 & 0.51 & 0.10 & 0.57 & 0.51 & 0.21 & 0.61 \\
Mistral-7B-Instruct & 0.10 & 0.56 & 0.50 & 0.60 & 0.61 & 0.49 & 0.47 & 0.45 & 0.10 & 0.42 & 0.41 & 0.15 & 0.58 \\
Gemma-7B-Instruct & 0.21 & 0.34 & 0.26 & 0.34 & 0.47 & 0.38 & 0.29 & 0.28 & -0.17 & 0.44 & 0.30 & -0.11 & 0.46 \\
Llama-2-7B-Chat & -0.03 & 0.33 & 0.33 & 0.40 & 0.46 & 0.33 & 0.17 & 0.18 & 0.01 & 0.27 & 0.26 & 0.11 & 0.41 \\
Salamandra-7B-Instruct & -0.09 & 0.40 & 0.35 & 0.39 & 0.28 & 0.25 & 0.17 & 0.20 & -0.21 & 0.28 & 0.12 & 0.06 & 0.41 \\
Leniachat-Gemma-2B & 0.04 & 0.15 & 0.10 & 0.03 & 0.23 & 0.07 & 0.13 & 0.03 & -0.01 & 0.15 & 0.09 & 0.05 & 0.07 \\
\midrule
SPANISH-noto &&&&&&&&&&&\\
\midrule
o3-mini & 0.68 & 0.41 & 0.49 & 0.41 & 0.27 & 0.25 & 0.58 & 0.08 & 0.28 & 0.26 & 0.28 & -0.15 & 0.51 \\
GPT-4o & 0.16 & 0.48 & 0.65 & 0.55 & 0.46 & 0.54 & 0.71 & 0.42 & 0.33 & 0.48 & 0.57 & -0.03 & 0.67 \\
GPT-4-Turbo & 0.09 & 0.51 & 0.69 & 0.61 & 0.42 & 0.63 & 0.62 & 0.48 & -0.19 & 0.44 & 0.48 & 0.03 & 0.57 \\
Deepseek-R1-70B & 0.07 & 0.28 & 0.33 & 0.38 & 0.46 & 0.29 & 0.20 & 0.16 & -0.19 & 0.30 & 0.25 & 0.23 & 0.35 \\
Claude-3.5-Sonnet & -0.05 & 0.34 & 0.55 & 0.53 & 0.46 & 0.37 & 0.31 & 0.27 & -0.03 & 0.33 & 0.37 & -0.01 & 0.42 \\
Gemma-2-27B-Instruct & -0.18 & 0.23 & 0.41 & 0.34 & 0.16 & 0.26 & 0.18 & 0.14 & -0.33 & 0.17 & 0.05 & -0.14 & 0.09 \\
Llama-3-70B-Instruct & -0.18 & 0.24 & 0.35 & 0.26 & 0.02 & 0.07 & 0.20 & -0.05 & -0.21 & 0.01 & -0.06 & -0.16 & 0.16 \\
Llama-3-8B-Instruct & -0.21 & 0.28 & 0.28 & 0.21 & 0.07 & 0.09 & 0.10 & 0.04 & -0.28 & 0.10 & -0.08 & -0.21 & -0.01 \\
Mixtral-8x7B-Instruct & -0.23 & 0.19 & 0.24 & 0.21 & -0.14 & -0.00 & 0.12 & -0.03 & -0.33 & -0.07 & -0.02 & -0.19 & 0.16 \\
Llama-2-7B-Chat & -0.26 & 0.33 & 0.32 & 0.27 & -0.14 & 0.01 & 0.06 & -0.07 & -0.33 & -0.15 & -0.09 & -0.23 & 0.16 \\
Mistral-7B-Instruct & -0.26 & 0.09 & 0.33 & 0.25 & -0.17 & -0.04 & 0.18 & 0.01 & -0.32 & -0.07 & -0.11 & -0.23 & 0.04 \\
Mixtral-8x22B-Instruct & -0.19 & -0.06 & 0.04 & -0.06 & -0.18 & -0.04 & -0.10 & -0.15 & -0.19 & -0.13 & -0.15 & -0.25 & -0.10 \\
Salamandra-7B-Instruct & -0.21 & -0.02 & 0.03 & -0.03 & -0.26 & -0.16 & -0.23 & -0.13 & -0.28 & -0.11 & -0.21 & -0.21 & -0.11 \\
GPT-3.5-Turbo & -0.29 & -0.14 & 0.00 & -0.11 & -0.26 & -0.28 & -0.28 & -0.21 & -0.33 & -0.24 & -0.26 & -0.30 & -0.29 \\
Gemma-7B-Instruct & -0.23 & -0.07 & -0.09 & -0.09 & -0.26 & -0.22 & -0.26 & -0.25 & -0.23 & -0.27 & -0.25 & -0.29 & -0.20 \\
Leniachat-Gemma-2B & -0.33 & -0.21 & -0.21 & -0.24 & -0.31 & -0.30 & -0.27 & -0.27 & -0.33 & -0.23 & -0.32 & -0.30 & -0.29 \\
\bottomrule

\bottomrule
\end{tabular}
}

\caption{Cohen's Kappa results on \textbf{MMLU} (\textbf{\textcolor{blue}{Humanities}}) by model and subject in English and Spanish, sorted by average.}
\label{tablammlu3}
\end{table*}

\begin{table*}[ht]
\centering
\resizebox{\linewidth}{!}{%
\renewcommand{\arraystretch}{0.8}
\begin{tabular}{lcccccccccccccccc}

 & \textbf{\begin{sideways}business ethics\end{sideways}} & \textbf{\begin{sideways}clinical knowledge\end{sideways}} & \textbf{\begin{sideways}college medicine\end{sideways}} & \textbf{\begin{sideways}global facts\end{sideways}} & \textbf{\begin{sideways}human aging\end{sideways}} & \textbf{\begin{sideways}human sexuality\end{sideways}} & \textbf{\begin{sideways}management\end{sideways}} & \textbf{\begin{sideways}marketing\end{sideways}} & \textbf{\begin{sideways}medical genetics\end{sideways}} & \textbf{\begin{sideways}miscellaneous\end{sideways}} & \textbf{\begin{sideways}nutrition\end{sideways}} & \textbf{\begin{sideways}professional accounting\end{sideways}} &
 \textbf{\begin{sideways}professional medicine\end{sideways}} &
 \textbf{\begin{sideways}professional psychology\end{sideways}} &
 \textbf{\begin{sideways}virology\end{sideways}} & \textbf{Average}\\
\toprule
ENGLISH-original  &&&&&&&&&&&&\\
\midrule
o3-mini & 0.73 & 0.87 & 0.84 & 0.49 & 0.77 & 0.91 & 0.87 & 0.89 & 0.99 & 0.93 & 0.91 & 0.87 & 0.95 & 0.84 & 0.42 & 0.85 \\
Claude-3.5-Sonnet & 0.76 & 0.91 & 0.78 & 0.70 & 0.80 & 0.90 & 0.90 & 0.95 & 0.97 & 0.96 & 0.89 & 0.78 & 0.94 & 0.89 & 0.37 & 0.82 \\
GPT-4o & 0.83 & 0.88 & 0.80 & 0.60 & 0.77 & 0.92 & 0.87 & 0.93 & 0.96 & 0.94 & 0.86 & 0.68 & 0.94 & 0.87 & 0.40 & 0.78 \\
GPT-4-Turbo & 0.75 & 0.82 & 0.70 & 0.42 & 0.81 & 0.89 & 0.83 & 0.92 & 0.95 & 0.93 & 0.83 & 0.62 & 0.89 & 0.83 & 0.38 & 0.74 \\
Llama-3-70B-Instruct & 0.76 & 0.80 & 0.70 & 0.45 & 0.77 & 0.85 & 0.86 & 0.86 & 0.90 & 0.88 & 0.82 & 0.49 & 0.83 & 0.78 & 0.34 & 0.71 \\
Gemma-2-27B-Instruct & 0.67 & 0.76 & 0.70 & 0.33 & 0.73 & 0.80 & 0.82 & 0.90 & 0.82 & 0.86 & 0.76 & 0.49 & 0.76 & 0.79 & 0.38 & 0.67 \\
Mixtral-8x22B-Instruct & 0.52 & 0.71 & 0.59 & 0.27 & 0.71 & 0.75 & 0.79 & 0.85 & 0.73 & 0.80 & 0.68 & 0.44 & 0.76 & 0.70 & 0.35 & 0.64 \\
Deepseek-R1-70B & 0.71 & 0.76 & 0.62 & 0.30 & 0.65 & 0.77 & 0.76 & 0.81 & 0.81 & 0.83 & 0.64 & 0.43 & 0.84 & 0.71 & 0.29 & 0.63 \\
Mixtral-8x7B-Instruct & 0.59 & 0.66 & 0.56 & 0.18 & 0.61 & 0.74 & 0.78 & 0.85 & 0.68 & 0.81 & 0.70 & 0.38 & 0.69 & 0.64 & 0.31 & 0.58 \\
GPT-3.5-Turbo & 0.60 & 0.67 & 0.51 & 0.29 & 0.63 & 0.75 & 0.75 & 0.88 & 0.68 & 0.84 & 0.63 & 0.35 & 0.71 & 0.61 & 0.32 & 0.54 \\
Llama-3-8B-Instruct & 0.55 & 0.67 & 0.50 & 0.10 & 0.64 & 0.66 & 0.80 & 0.87 & 0.71 & 0.77 & 0.66 & 0.35 & 0.66 & 0.57 & 0.33 & 0.52 \\
Mistral-7B-Instruct & 0.48 & 0.52 & 0.42 & 0.16 & 0.53 & 0.60 & 0.69 & 0.84 & 0.49 & 0.71 & 0.56 & 0.21 & 0.49 & 0.45 & 0.26 & 0.46 \\
Gemma-7B-Instruct & 0.36 & 0.42 & 0.29 & 0.06 & 0.53 & 0.45 & 0.52 & 0.72 & 0.41 & 0.58 & 0.39 & 0.07 & 0.27 & 0.35 & 0.20 & 0.32 \\
Llama-2-7B-Chat & 0.15 & 0.29 & 0.20 & 0.10 & 0.27 & 0.30 & 0.23 & 0.52 & 0.16 & 0.42 & 0.26 & 0.07 & 0.11 & 0.19 & 0.13 & 0.20 \\
Salamandra-7B-Instruct & 0.39 & 0.22 & 0.15 & -0.08 & 0.27 & 0.18 & 0.32 & 0.44 & 0.18 & 0.37 & 0.00 & 0.04 & 0.11 & 0.21 & 0.14 & 0.17 \\
Leniachat-Gemma-2B & 0.15 & 0.15 & 0.11 & -0.02 & 0.12 & -0.02 & 0.18 & 0.27 & 0.13 & 0.26 & 0.13 & 0.03 & 0.08 & 0.08 & 0.10 & 0.10 \\
\midrule
ENGLISH-noto &&&&&&&&&&&&\\
\midrule
o3-mini & 0.21 & 0.62 & 0.58 & 0.04 & 0.18 & 0.44 & 0.40 & 0.44 & 0.82 & 0.76 & 0.55 & 0.50 & 0.82 & 0.34 & 0.14 & 0.54 \\
Deepseek-R1-70B & 0.20 & 0.59 & 0.52 & 0.58 & 0.29 & 0.32 & 0.49 & 0.19 & 0.51 & 0.73 & 0.34 & 0.50 & 0.60 & 0.34 & 0.19 & 0.46 \\
GPT-4o & 0.32 & 0.61 & 0.46 & 0.23 & 0.37 & 0.58 & 0.46 & 0.57 & 0.71 & 0.79 & 0.51 & 0.21 & 0.57 & 0.43 & 0.19 & 0.45 \\
GPT-4-Turbo & 0.19 & 0.46 & 0.26 & 0.21 & 0.36 & 0.46 & 0.46 & 0.43 & 0.52 & 0.76 & 0.36 & 0.11 & 0.30 & 0.27 & 0.14 & 0.34 \\
Claude-3.5-Sonnet & -0.12 & 0.30 & 0.12 & -0.12 & 0.11 & 0.21 & 0.18 & 0.21 & 0.45 & 0.52 & 0.19 & 0.07 & 0.24 & 0.10 & -0.01 & 0.20 \\
Gemma-2-27B-Instruct & -0.03 & 0.17 & 0.08 & -0.19 & 0.10 & 0.19 & 0.16 & 0.19 & 0.27 & 0.32 & 0.11 & -0.11 & 0.13 & 0.09 & -0.07 & 0.09 \\
Llama-3-70B-Instruct & -0.13 & 0.17 & 0.07 & -0.13 & -0.01 & 0.14 & 0.03 & 0.11 & 0.42 & 0.38 & 0.11 & -0.09 & 0.37 & 0.01 & -0.01 & 0.06 \\
Mixtral-8x7B-Instruct & -0.12 & 0.03 & -0.02 & 0.03 & 0.05 & 0.12 & 0.16 & 0.17 & 0.23 & 0.46 & 0.00 & -0.12 & 0.05 & -0.05 & -0.07 & 0.05 \\
Llama-2-7B-Chat & -0.12 & 0.02 & -0.01 & 0.10 & 0.22 & 0.14 & 0.27 & 0.07 & -0.06 & 0.26 & -0.05 & -0.08 & 0.21 & 0.06 & 0.06 & 0.05 \\
Llama-3-8B-Instruct & -0.19 & 0.12 & 0.04 & -0.20 & -0.03 & 0.04 & 0.02 & 0.08 & 0.24 & 0.23 & 0.06 & -0.08 & 0.19 & -0.02 & -0.09 & 0.02 \\
Mistral-7B-Instruct & -0.19 & -0.10 & -0.11 & -0.21 & -0.09 & -0.03 & 0.03 & 0.04 & 0.07 & 0.13 & -0.12 & -0.24 & -0.19 & -0.13 & -0.17 & -0.09 \\
Mixtral-8x22B-Instruct & -0.15 & -0.04 & -0.11 & -0.23 & -0.18 & -0.06 & -0.01 & -0.10 & 0.01 & -0.00 & -0.12 & -0.25 & -0.05 & -0.16 & -0.24 & -0.12 \\
Salamandra-7B-Instruct & -0.13 & -0.11 & -0.11 & -0.20 & -0.22 & -0.21 & -0.25 & -0.21 & -0.14 & -0.15 & -0.02 & -0.19 & -0.16 & -0.21 & -0.23 & -0.18 \\
GPT-3.5-Turbo & -0.31 & -0.24 & -0.18 & -0.31 & -0.19 & -0.20 & -0.14 & -0.17 & -0.20 & -0.03 & -0.21 & -0.27 & -0.29 & -0.26 & -0.22 & -0.21 \\
Gemma-7B-Instruct & -0.25 & -0.20 & -0.24 & -0.27 & -0.22 & -0.20 & -0.05 & -0.15 & -0.20 & -0.15 & -0.26 & -0.25 & -0.22 & -0.27 & -0.27 & -0.24 \\
Leniachat-Gemma-2B & -0.27 & -0.32 & -0.31 & -0.28 & -0.31 & -0.32 & -0.32 & -0.32 & -0.32 & -0.30 & -0.30 & -0.31 & -0.16 & -0.31 & -0.29 & -0.29 \\
\midrule
SPANISH-original  &&&&&&&&&&&&\\
\midrule
o3-mini & 0.75 & 0.80 & 0.80 & 0.52 & 0.76 & 0.88 & 0.83 & 0.88 & 0.97 & 0.93 & 0.84 & 0.79 & 0.94 & 0.80 & 0.38 & 0.82 \\
Claude-3.5-Sonnet & 0.79 & 0.86 & 0.74 & 0.66 & 0.76 & 0.90 & 0.88 & 0.94 & 0.95 & 0.94 & 0.87 & 0.71 & 0.93 & 0.86 & 0.43 & 0.80 \\
GPT-4o & 0.75 & 0.83 & 0.76 & 0.46 & 0.70 & 0.89 & 0.84 & 0.93 & 0.93 & 0.94 & 0.83 & 0.63 & 0.92 & 0.85 & 0.37 & 0.76 \\
GPT-4-Turbo & 0.61 & 0.79 & 0.69 & 0.31 & 0.79 & 0.84 & 0.83 & 0.90 & 0.90 & 0.90 & 0.76 & 0.50 & 0.86 & 0.80 & 0.35 & 0.70 \\
Llama-3-70B-Instruct & 0.71 & 0.75 & 0.62 & 0.38 & 0.69 & 0.79 & 0.75 & 0.85 & 0.78 & 0.84 & 0.71 & 0.37 & 0.78 & 0.67 & 0.32 & 0.65 \\
Gemma-2-27B-Instruct & 0.60 & 0.69 & 0.65 & 0.31 & 0.69 & 0.76 & 0.76 & 0.88 & 0.77 & 0.83 & 0.72 & 0.36 & 0.68 & 0.69 & 0.26 & 0.63 \\
Deepseek-R1-70B & 0.72 & 0.67 & 0.60 & 0.35 & 0.66 & 0.80 & 0.73 & 0.82 & 0.78 & 0.82 & 0.71 & 0.34 & 0.75 & 0.69 & 0.30 & 0.63 \\
Mixtral-8x22B-Instruct & 0.55 & 0.64 & 0.58 & 0.16 & 0.59 & 0.72 & 0.71 & 0.84 & 0.62 & 0.75 & 0.65 & 0.35 & 0.64 & 0.60 & 0.29 & 0.56 \\
Mixtral-8x7B-Instruct & 0.60 & 0.57 & 0.51 & 0.15 & 0.53 & 0.67 & 0.66 & 0.75 & 0.64 & 0.78 & 0.64 & 0.25 & 0.61 & 0.54 & 0.28 & 0.52 \\
GPT-3.5-Turbo & 0.47 & 0.57 & 0.47 & 0.18 & 0.57 & 0.65 & 0.62 & 0.78 & 0.57 & 0.78 & 0.52 & 0.22 & 0.53 & 0.51 & 0.28 & 0.46 \\
Llama-3-8B-Instruct & 0.49 & 0.47 & 0.39 & -0.01 & 0.50 & 0.56 & 0.65 & 0.68 & 0.48 & 0.65 & 0.49 & 0.11 & 0.46 & 0.47 & 0.24 & 0.41 \\
Mistral-7B-Instruct & 0.43 & 0.39 & 0.27 & 0.12 & 0.43 & 0.44 & 0.53 & 0.72 & 0.35 & 0.60 & 0.41 & 0.19 & 0.29 & 0.34 & 0.19 & 0.35 \\
Gemma-7B-Instruct & 0.27 & 0.30 & 0.22 & 0.11 & 0.43 & 0.38 & 0.45 & 0.63 & 0.29 & 0.47 & 0.29 & 0.07 & 0.12 & 0.24 & 0.15 & 0.26 \\
Llama-2-7B-Chat & 0.16 & 0.19 & 0.15 & 0.19 & 0.26 & 0.14 & 0.18 & 0.45 & 0.16 & 0.38 & 0.21 & 0.08 & 0.05 & 0.18 & 0.23 & 0.17 \\
Salamandra-7B-Instruct & 0.29 & 0.16 & 0.15 & -0.05 & 0.21 & 0.24 & 0.22 & 0.41 & 0.11 & 0.36 & 0.17 & 0.02 & 0.13 & 0.20 & 0.08 & 0.14 \\
Leniachat-Gemma-2B & 0.00 & 0.01 & 0.03 & 0.03 & 0.04 & 0.02 & 0.16 & 0.15 & 0.09 & 0.14 & 0.03 & 0.03 & 0.10 & 0.09 & 0.08 & 0.07 \\
\midrule
SPANISH-noto &&&&&&&&&&&&\\
\midrule
o3-mini & 0.15 & 0.53 & 0.51 & -0.01 & 0.07 & 0.41 & 0.20 & 0.33 & 0.77 & 0.72 & 0.44 & 0.40 & 0.70 & 0.25 & 0.14 & 0.47 \\
GPT-4o & 0.29 & 0.54 & 0.37 & 0.34 & 0.32 & 0.51 & 0.42 & 0.48 & 0.66 & 0.76 & 0.45 & 0.09 & 0.55 & 0.35 & 0.18 & 0.41 \\
GPT-4-Turbo & 0.23 & 0.43 & 0.30 & 0.35 & 0.46 & 0.41 & 0.49 & 0.43 & 0.48 & 0.75 & 0.33 & 0.16 & 0.37 & 0.28 & 0.21 & 0.37 \\
Deepseek-R1-70B & -0.05 & 0.35 & 0.29 & 0.31 & 0.11 & 0.32 & 0.19 & 0.15 & 0.34 & 0.53 & 0.24 & 0.25 & 0.22 & 0.03 & 0.07 & 0.25 \\
Claude-3.5-Sonnet & -0.01 & 0.31 & 0.18 & -0.09 & 0.12 & 0.34 & 0.22 & 0.24 & 0.48 & 0.61 & 0.26 & 0.05 & 0.33 & 0.18 & 0.06 & 0.24 \\
Gemma-2-27B-Instruct & -0.12 & 0.13 & 0.00 & -0.20 & 0.02 & 0.16 & 0.18 & 0.11 & 0.12 & 0.22 & 0.03 & -0.13 & 0.04 & 0.02 & -0.08 & 0.04 \\
Llama-3-70B-Instruct & -0.15 & 0.09 & 0.00 & -0.16 & -0.04 & 0.03 & 0.10 & 0.10 & 0.24 & 0.24 & 0.03 & -0.17 & 0.06 & -0.08 & -0.09 & 0.00 \\
Llama-3-8B-Instruct & -0.28 & 0.01 & -0.15 & -0.14 & -0.08 & 0.03 & -0.01 & 0.06 & -0.03 & 0.17 & -0.06 & -0.12 & -0.04 & -0.07 & -0.10 & -0.03 \\
Mixtral-8x7B-Instruct & -0.15 & -0.10 & -0.11 & 0.07 & -0.09 & -0.07 & 0.15 & 0.04 & 0.02 & 0.24 & -0.11 & -0.16 & -0.09 & -0.11 & -0.14 & -0.04 \\
Llama-2-7B-Chat & -0.16 & -0.07 & -0.16 & 0.08 & 0.02 & 0.03 & 0.14 & 0.02 & -0.15 & 0.15 & -0.01 & -0.07 & 0.07 & -0.11 & 0.00 & -0.05 \\
Mistral-7B-Instruct & -0.21 & -0.11 & -0.15 & -0.23 & -0.08 & -0.05 & 0.03 & 0.05 & 0.01 & 0.09 & -0.10 & -0.18 & -0.09 & -0.13 & -0.16 & -0.07 \\
Mixtral-8x22B-Instruct & -0.20 & -0.04 & -0.07 & -0.25 & -0.17 & -0.16 & -0.03 & -0.09 & -0.07 & 0.06 & -0.13 & -0.23 & -0.11 & -0.16 & -0.21 & -0.12 \\
Salamandra-7B-Instruct & -0.17 & -0.10 & -0.16 & -0.19 & -0.19 & -0.19 & -0.11 & -0.11 & -0.15 & -0.15 & -0.20 & -0.19 & -0.09 & -0.18 & -0.20 & -0.15 \\
GPT-3.5-Turbo & -0.27 & -0.21 & -0.22 & -0.25 & -0.19 & -0.27 & -0.18 & -0.19 & -0.21 & -0.11 & -0.26 & -0.29 & -0.25 & -0.27 & -0.27 & -0.23 \\
Gemma-7B-Instruct & -0.23 & -0.16 & -0.22 & -0.29 & -0.21 & -0.19 & -0.03 & -0.21 & -0.22 & -0.15 & -0.24 & -0.25 & -0.21 & -0.26 & -0.25 & -0.23 \\
Leniachat-Gemma-2B & -0.33 & -0.30 & -0.30 & -0.32 & -0.30 & -0.30 & -0.33 & -0.29 & -0.29 & -0.29 & -0.31 & -0.28 & -0.27 & -0.28 & -0.32 & -0.30 \\
\bottomrule
\end{tabular}
}

\caption{Cohen's Kappa results on \textbf{MMLU} (\textbf{\textcolor{blue}{Other categories}}) by model and subject in English and Spanish, sorted by average.}
\label{tablammlu4}
\end{table*}

\end{document}